%% file: neurips_2025.tex
\documentclass{article}

\usepackage[preprint]{neurips_2025}
\usepackage{enumitem}

\usepackage[utf8]{inputenc} 
\usepackage[T1]{fontenc}    
\usepackage{hyperref}       
\usepackage{url}            
\usepackage{booktabs}       
\usepackage{amsfonts}       
\usepackage{nicefrac}       
\usepackage{microtype}      
\usepackage{xcolor}         
\usepackage{times}
\usepackage{latexsym}
\usepackage[T1]{fontenc}
\usepackage{svg}
\usepackage{wrapfig}
\usepackage{booktabs}

\usepackage[utf8]{inputenc}
\usepackage{amsthm}

\usepackage{microtype}

\usepackage{inconsolata}

\usepackage{graphicx}
\usepackage{multirow}
\newtheorem{theorem}{Theorem}
\usepackage{amsthm}

\usepackage{tabularx}
\usepackage{mathptmx}
\newcommand{\Tau}{T}
\usepackage{amsmath}
\usepackage{amssymb}
\usepackage{algorithm}
\usepackage{algorithmic}
\usepackage{subcaption}
\usepackage{caption}
\usepackage{booktabs}  
\usepackage{adjustbox}

\newcommand{\ModelName}{POSTRA}
\newcommand{\parabold}[1]{\vspace{2pt}\noindent\textbf{#1}}
\usepackage[utf8]{inputenc}
\usepackage{textcomp}
\title{Towards Foundation Model on Temporal Knowledge Graph Reasoning}

\author{%
Jiaxin Pan$^{1}$ \quad Mojtaba Nayyeri $^{1}$ \quad Osama Mohammed$^1$ \quad Daniel Hernández$^1$ \\ \textbf{Rongchuan Zhang}$^1$ \quad
\textbf{Cheng Cheng}$^1$ \quad \textbf{Steffen Staab}$^{1,2}$\\
$^1$University of Stuttgart \quad $^2$University of Southampton \\
\texttt{\{jiaxin.pan,mojtaba.nayyeri,osama.mohammed,daniel.hernandez\}@ki.uni-stuttgart.de}\\
\texttt{\{st191486, st180913\}@stud.uni-stuttgart.de}\\
\texttt{\{steffen.staab\}@ki.uni-stuttgart.de}
}

\input{latex/macros}

\begin{document}

\maketitle

\begin{abstract}

Temporal Knowledge Graphs (TKGs) store temporal facts with quadruple formats $(s, p, o, \tau)$. Existing Temporal Knowledge Graph Embedding (TKGE) models perform link prediction tasks in transductive or semi-inductive settings, which means the entities, relations, and temporal information in the test graph are fully or partially observed during training. Such reliance on seen elements during inference limits the models’ ability to transfer to new domains and generalize to real-world scenarios. A central limitation is the difficulty in learning representations for entities, relations, and timestamps that are transferable and not tied to dataset-specific vocabularies. To overcome these limitations, we introduce the first fully-inductive approach to temporal knowledge graph link prediction. Our model employs sinusoidal positional encodings to capture fine-grained temporal patterns and generates adaptive entity and relation representations using message passing conditioned on both local and global temporal contexts. Our model design is agnostic to temporal granularity and time span, effectively addressing temporal discrepancies across TKGs and facilitating time-aware structural information transfer.
As a pretrained, scalable, and transferable model, \ModelName{} demonstrates strong zero-shot performance on unseen temporal knowledge graphs, effectively generalizing to novel entities, relations, and timestamps. Extensive theoretical analysis and empirical results show that a single pretrained model can improve zero-shot performance on various inductive temporal reasoning scenarios, marking a significant step toward a foundation model for temporal KGs.

\end{abstract}

\input{latex/sec_intro}

\input{latex/sec_task_formation}

\input{latex/sec_related_work}

\input{latex/sec_methodology}

\input{latex/sec_experiment_setup}

\input{latex/sec_result}

\section{Conclusion}
We introduce \ModelName{}, which advances beyond existing transductive or semi-inductive approaches to support fully inductive inference. \ModelName{} exhibits strong universal transferability across diverse time spans, temporal granularities, domains, and prediction tasks, overcoming limitations of prior models that depended on dataset-specific entity, relation, and timestamp embeddings. Extensive evaluations across multiple scenarios show that \ModelName{} consistently outperforms existing state-of-the-art models in zero-shot settings. In future work, we aim to extend \ModelName{} to support time prediction tasks and explore richer temporal representations.

\section*{Acknowledgment}
This work has received funding from the CHIPS Joint Undertaking (JU) under grant agreement No. 101140087 (SMARTY). The JU receives support from the European Union’s Horizon Europe research and innovation programme. Furthermore, on national level this work is supported by the German Federal Ministry of Education and Research (BMBF) under the sub-project with the funding number 16MEE0444.
The authors gratefully acknowledge the computing time provided on the high-performance computer HoreKa by the National High-Performance Computing Center at KIT (NHR@KIT). This center is jointly supported by the Federal Ministry of Education and Research and the Ministry of Science, Research and the Arts of Baden-Württemberg, as part of the National High-Performance Computing (NHR) joint funding program (https://www.nhr-verein.de/en/our-partners). HoreKa is partly funded by the German Research Foundation (DFG).

\newpage

\bibliographystyle{plainnat}
\bibliography{latex/neurips_2025}



\newpage

\appendix

\input{latex/sec_appendix}

\end{document}

%% file: latex/macros.tex
\newcommand{\Glocal}{G_{\mathit{local}}}

%% file: latex/sec_intro.tex
\section{Introduction}
Temporal Knowledge Graphs (TKGs) extend static knowledge graphs with temporal information. In TKGs, temporal facts are represented by the quadruple format $(s, p, o, \tau)$, where $s$, $p$, $o$ and $\tau$ denote head entity, relation name, tail entity and timestamp, respectively. Temporal Knowledge Graphs (TKGs) have been researched in various fields, including question answering \citep{saxena-etal-2021-question}, data integration \citep{10.5555/AAI30283568}, and entity alignment \citep{cai2024surveytemporalknowledgegraph}.

\begin{wrapfigure}{l}{0.65\textwidth}
    \centering
    \includegraphics[width=0.65\textwidth]{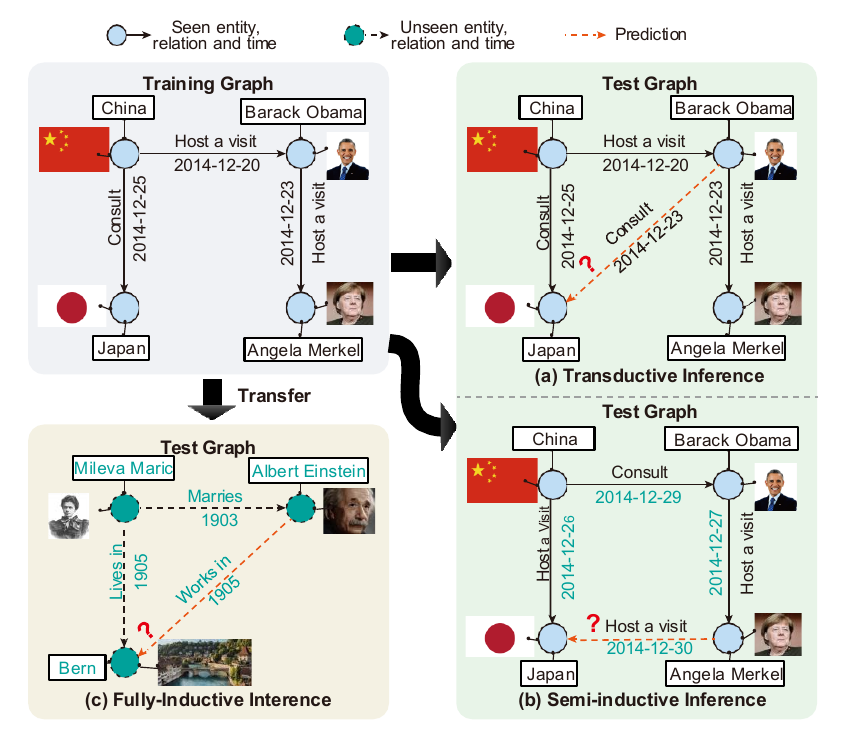}
    
    \caption{
 Subfigure (a) shows the \textit{transductive} setting, where the test graph contains only entities, relation names, and timestamps that have already been seen during training.  
Subfigure (b) demonstrates the \textit{semi-inductive} setting, where all entities and relation names in the test graph are present in the training graph, but the timestamps (e.g., \texttt{2014-12-30}) are strictly later than those observed during training.  
Subfigure (c) illustrates the proposed \textit{fully-inductive} setting, where the test graph includes  unseen entities (e.g., \texttt{Mileva Marić}), relation names (e.g., \texttt{lives in}), and timestamps (e.g., \texttt{1905}).  
    }
    \label{fig:inference_example}
\end{wrapfigure}

Link prediction is a crucial task in temporal knowledge graphs (TKGs), which involves predicting missing entities in temporal facts. Given a query of the form (?, $p$, $o$, $\tau$) or ($s$, $p$, ?, $\tau$), the objective is to infer
 head or tail entity at a specific given timestamp $\tau$. This task is essential for learning effective temporal embeddings that support various downstream applications. However, existing temporal knowledge graph embedding methods that follow transductive or semi-inductive settings suffer from transferability issues. As shown in Fig. \ref{fig:inference_example} (a), the transductive setting (temporal knowledge graph interpolation) shares all the entities, relations, and timestamps information in both training and testing. Fig. \ref{fig:inference_example} (b) depicts the  semi-inductive setting (temporal knowledge graph extrapolation or forecasting), where the timestamp information in the test graph is not available during training. 
However, the entities and/or relations remain consistent between the training and test graphs.


As existing TKG embedding models rely on the entity, relation vocabularies and used timestamps to train specific embeddings for prediction, they need to be retrained whenever a new TKG is introduced. This limitation hinders their generalization to real-world scenarios when new entities, relations, and timestamps emerge (Fig. \ref{fig:inference_example} c).
To overcome the generalization limitation in TKG models, we present a fully inductive TKG learning framework, \textbf{\ModelName{}}, a model for \underline{pos}itional \underline{t}ime-aware, and \underline{tra}nsferable inductive TKG representations. Specifically, we address two key challenges:
1) Temporal discrepancies across TKGs and
2) Temporal-aware structural information transfer across TKGs.

\begin{wrapfigure}{r}{0.5\textwidth}
    \centering
    \includegraphics[width=0.48\textwidth]{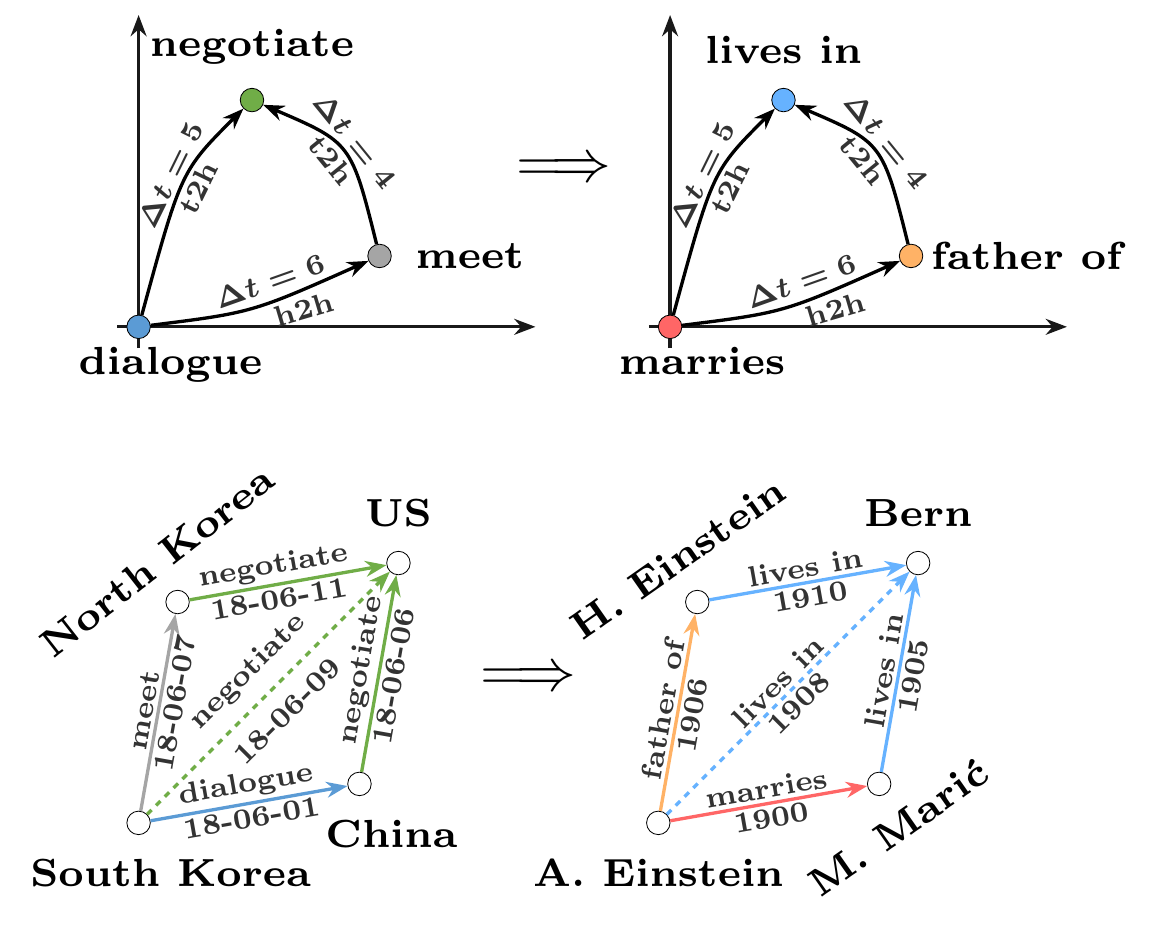}
    \caption{
    The bottom TKGs depict temporal knowledge graphs from different domains. The upper relation graphs show relative relation representations learned via fundamental interaction patterns (See Section \ref{sec:relation_learning}) and the relative temporal ordering between corresponding facts ($\Delta t$) (See Section \ref{sec:temporal_embedding}) which are transferrable across TKGs. More details are shown in Section \ref{sec:relative_representation} in Appendix.
    }
    \label{fig:temporal_relation_transfer}
\end{wrapfigure}

Starting with the first challenge, different TKGs utilize varying time units (e.g., minute/day/month) and cover diverse temporal spans (e.g., 1 month/100 years) depending on the frequency of temporal facts. 
Existing TKG embedding models\citep{tcomplexlacroix2020tensor, tltcomplexzhang2022along} often rely on dataset-specific trained time embeddings, which limits their ability to transfer learned temporal information across TKGs with different granularities and time spans. 
To address this, we leverage the transferability of relative temporal ordering between connected facts along temporal paths. We organize a TKG as a sequence of temporal snapshots, where each snapshot contains all facts sharing the same timestamp. We focus on the relative ordering of quadruples in different snapshots, rather than the specific timestamp values, and encode this ordering using sinusoidal positional encodings as time series embeddings. For example, in Figure \ref{fig:temporal_relation_transfer}, although the temporal facts in two TKGs below have different timestamps, the relative temporal ordering difference between connected temporal facts is the same as shown in the relation graphs above. Therefore, this representation captures the temporal dependencies between facts in a way that is invariant to the underlying time units or time spans and transferable to new domains, effectively bypassing the temporal discrepancies challenge. By integrating these relative temporal signals into the message passing process of relation encoder and quadruple encoder, the model can learn temporal patterns without relying on dataset-specific granularity or time span.

To address the second challenge of temporal-aware structural information transfer, we adopt four core relation interaction types—head-head, head-tail, tail-head, and tail-tail—that are independent of dataset-specific relation names\citep{galkintowards,lee2023ingram} and introduce a temporal-aware message passing mechanism. By constructing entity and relation embeddings as functions conditioned on these intrinsic interactions, a pre-trained model can inductively generalize to any knowledge graph through the message-passing process, even when encountering unseen entities and relations. To incorporate temporal dynamics, \ModelName{} injects temporal positional embeddings into the message-passing process, enabling the resulting entity and relation representations to capture time-dependent patterns, as illustrated in Figure~\ref{fig:temporal_relation_transfer}. Moreover, to differentiate between queries that share the same entities and relation names but occur at different times (e.g.,($s$, $p$, $o$, $\tau$) v.s. ($s$, $p$, $o$, $\tau'$)), \ModelName{} adopts a dual training strategy that separately learns local and global temporal structures. Global training aggregates information across the entire training graph, capturing long-term temporal dependencies, while local training restricts updates to a fixed temporal window around each target query, focusing on short-term temporal patterns. By combining these two complementary signals, \ModelName{} effectively adapts to TKGs with diverse temporal dynamics and varying time spans.

 Different from existing TKGE models, our model does not rely on any dataset-specific entity, relation, or time embeddings (dataset-specific vocabulary embedding) and can generalize to any new TKG. While this allows the pretrained model to perform zero-shot inference across diverse TKGs, it also remains compatible with transductive and semi-inductive TKG link prediction tasks, providing flexibility across different task settings. 
 Our model’s strong performance in transferability and cross‑domain generalization across diverse datasets underscores its effectiveness and marks a key advance toward a foundation model for temporal knowledge graphs.

%% file: latex/sec_task_formation.tex
\section{Task Formulation}

Let $V$ be a finite set of \emph{entities}, $R$ be a finite set of \emph{relation names}, and $\Tau = (\tau_i)_{i=1}^{|\Tau|}$ be a finite ordered set of \emph{timestamps}. 
A \emph{temporal knowledge graph} is a quad $G = (V, R, \Tau, Q)$, where $Q\subseteq V\times R \times V \times \Tau$.
The elements of $Q$ are called \emph{temporal facts} in quadruple form.
Given a positive natural number $i$, with $i \leq |\Tau|$, the \emph{\(i\)-snapshot} of \(G\), denoted \(G_i\), is the subgraph of \(G\) that contains all temporal facts whose timestamp is the $i$-th element of $\Tau$. We write $\tau_i$ for the timestamp at snapshot $G_i$.
Intuitively, a temporal fact describes a relationship between two entities at a determined timestamp, and a snapshot describes all relationships that occur simultaneously at a determined timestamp.



\newcommand{\GG}{{G}}
\newcommand{\VV}{{V}}
\newcommand{\RR}{{R}}
\newcommand{\QQ}{\{Q\}}
\newcommand{\Gtrain}{{G}_{\mathit{train}}}
\newcommand{\Vtrain}{{V}_{\mathit{train}}}
\newcommand{\Rtrain}{{R}_{\mathit{train}}}
\newcommand{\Ttrain}{\Tau_{\mathit{train}}}
\newcommand{\Qtrain}{{Q}_{\mathit{train}}}
\newcommand{\Gobs}{{G}_{\textsc{o}}}
\newcommand{\Vobs}{{V}_{\textsc{o}}}
\newcommand{\Robs}{{R}_{\textsc{o}}}
\newcommand{\Tobs}{\Tau_{\textsc{o}}}
\newcommand{\Qobs}{{Q}_{\textsc{o}}}
\newcommand{\Gvalid}{{G}_{\mathit{valid}}}
\newcommand{\Vvalid}{{V}_{\mathit{valid}}}
\newcommand{\Rvalid}{{R}_{\mathit{valid}}}
\newcommand{\Tvalid}{\Tau_{\mathit{valid}}}
\newcommand{\Qvalid}{{Q}_{\mathit{valid}}}
\newcommand{\Gtest}{{G}_{\mathit{test}}}
\newcommand{\Vtest}{{V}_{\mathit{test}}}
\newcommand{\Rtest}{{R}_{\mathit{test}}}
\newcommand{\Ttest}{\Tau_{\mathit{test}}}
\newcommand{\Qtest}{{Q}_{\mathit{test}}}
\newcommand{\Ginf}{{G}_{\mathit{inf}}}
\newcommand{\Vinf}{{V}_{\mathit{inf}}}
\newcommand{\Rinf}{{R}_{\mathit{inf}}}
\newcommand{\Tinf}{\Tau_{\mathit{inf}}}
\newcommand{\Qinf}{{Q}_{\mathit{inf}}}
\newcommand{\txt}{\Sigma^*}
\newcommand{\labl}{{l}}

\textbf{Link Prediction.} 
We aim at link prediction, the fundamental task for TKG reasoning which predicts missing entities of queries $(s, p, ?, \tau)$ or $(?, p, o, \tau)$. 
For this task, we assume: (i) one \emph{training graph} $\Gtrain=(\Vtrain,\Rtrain,\Ttrain,\Qtrain)$. (ii) one \emph{inference graph} $\Ginf=(\Vinf,\Rinf,\Tinf,\Qinf)$. We divide $\Qinf$ into three pairwise disjoint sets, such that $\Qinf=\Qobs \cup \Qvalid \cup \Qtest$. $\Gobs=(\Vobs,\Robs,\Tobs,\Qobs)$, $\Gvalid=(\Vvalid,\Rvalid,\Tvalid,\Qvalid)$, $\Gtest=(\Vtest,\Rtest,\Ttest,\Qtest)$ refers to the \emph{observed graph}, \emph{validation graph} and \emph{test graph}, respectively.


At training time, we use $\Gtrain$ to train the model to predict $\Qtrain$. At inference, we use $\Qobs$ to compute the embeddings and tune hyperparameters based on the model's performance on $\Qvalid$. Finally, we test the model's performance on $\Qtest$.  

\textbf{Transductive Inference} 
 This setting requires that all entities, relation names, and timestamps are known during the training process. The inference graph contains only known entities, relation names, and timestamps (see Fig. \ref{fig:inference_example}-a). 
In this setting, we have $\Vtrain = \Vinf, \Rtrain=\Rinf$ and $\Ttrain=\Tinf$. 
This implies that the entity set, relation name set, and timestamp set remain identical in both the training and test processes.


\textbf{Semi-Inductive Inference}
In this setting, the entity and relation name sets are shared between the training and test graphs: $\Vtrain = \Vinf, \Rtrain=\Rinf$. Similarly, times are shared between the training and the observed graphs: $\Tobs = \Ttrain$. However, this task enforces the constraint: $\forall \tau_i \in \Ttrain, \forall \tau_j \in \Tvalid, \forall \tau_k \in \Ttest, \tau_i < \tau_j < \tau_k$ (see Fig. \ref{fig:inference_example}-b).
This ensures that all timestamps in the validation set are strictly later than those in the training set, and all timestamps in the test set are strictly later than those in the validation set, thereby meeting the requirement for predicting future temporal facts. It enables semi-inductive inference on future timestamps.

However, the two existing inference types do not support cross-dataset inductive inference, as the entity and relation sets remain shared across datasets.

\textbf{Fully-Inductive Inference}
We propose the more challenging task of fully inductive temporal knowledge graph inference, in which the inference graph contains entities, relation names, and timestamps that never appear in the training data (see Fig. \ref{fig:inference_example}-c).
This formally means $\Vtrain \cap \Vinf = \emptyset, \Rtrain \cap \Rinf = \emptyset, \Ttrain \cap \Tinf = \emptyset$. 
Unlike temporal extrapolation inference, this setting does not impose constraints on the sequence of timestamps between the training and inference graphs. In this setting, the pretrained model utilizes weights learned from the training graph to generate adaptive embeddings for the inference graph (with a totally new graph vocabulary).
Crucially, a genuine foundation model for temporal knowledge graphs must work in this fully inductive setting and achieve maximal transferability of temporal structural patterns. 
\textbf{To our knowledge, no existing structural temporal knowledge graph learning model works entirely in a fully inductive setting— a gap this paper aims to close as an important step toward foundation models for temporal knowledge graphs.}



\begin{figure}[!t]
    \centering
    \includegraphics[width=\textwidth]{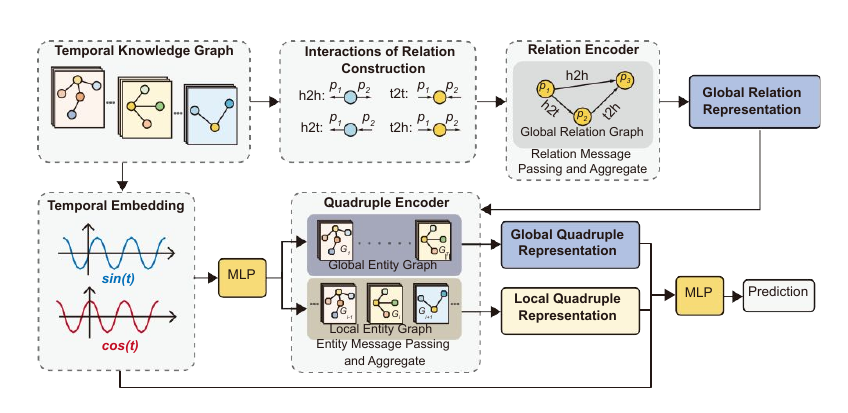}
    \caption{
 The overall architecture of POSTRA. The model first constructs a Global Relation Graph by treating relations as nodes and the four types of interactions of relations as edges. The Relation Encoder learns global relation representations on the global relation graph through meassage-passing, which are then used in downstream quadruple encoding(See Section \ref{sec:relation_learning}). Temporal Embedding encodes timestamps' relative ordering in the TKG by sine and cosine functions(See Section \ref{sec:temporal_embedding}). The Quadruple Encoder processes entity representations from both Global Entity Graphs and Local Entity Graphs (see Section \ref{sec:entity_learning}), generating Global and Local Quadruple Representations. Finally, these representations, along with temporal embeddings, are fused through an MLP to produce the final Prediction.
    }
    \label{fig:model_overview}
\end{figure}

%% file: latex/sec_related_work.tex
\section{Related Work}
\subsection{Temporal Knowledge Graph Embedding}
\textbf{Transductive Models} TTransE \citep{leblay2018deriving} and TA-DistMult \citep{garcia-duran-etal-2018-learning} are among the earliest models to incorporate temporal information into score functions by treating time as an additional element. 
T(NT)ComplEx \citep{tcomplexlacroix2020tensor} formulates temporal knowledge graph completion as a fourth-order tensor completion problem using semantic matching. Later work TLT-KGE \citep{tltcomplexzhang2022along} extends it to quaternion space and exchanges information through quaternion operations. HGE \citep{pan2024hge} embeds temporal knowledge graphs into a product space of suitable manifolds and adopts an attention mechanism that captures the similarities of structural patterns and manifolds.
This line of work lacks generalization to unseen TKGs.

\noindent \textbf{Temporal Extrapolation Models} Most existing models utilize graph neural network architectures with various updating strategies to predict future facts \citep{li2022complex,li2022tirgn,sun2021timetraveler,liang2023learn}. For improved explainability, TLogic \citep{liu2022tlogic} extracts temporal rules using temporal random walks, while xERTE \citep{han2020explainable} performs predictions based on explainable subgraphs.

Additionally, TPAR \citep{chen2024unified} is a model capable of handling both transductive and temporal extrapolation tasks. It updates entity representations using Bellman-Ford-based recursive encoding on temporal paths. However, TPAR is unable to perform inductive inference where new entities, relations, and timestamps emerge.

\subsection{Inductive Learning on KGs}
\textbf{Structural Inductive Learning} Early inductive models, such as NBFNet \citep{zhu2021neural}, Grail \citep{teru2020inductive}, INDIGO \citep{liu2021indigo}, and Morse \citep{chen2022meta}, learn relational patterns in knowledge graphs and can handle unseen entities. However, they require that the relations in the training and inference graphs remain the same. Later works, such as INGRAM \citep{lee2023ingram}, TARGI \citep{ding2025towards} and ULTRA \citep{galkintowards}, construct relation graphs where relations are treated as nodes and interactions between relations as edges. The relations are learned through similar interaction structures utilized for entity representation learning. This enables structural information sharing across training and inference graphs, allowing them to predict unseen entities and relations. However, none of these models can infer temporal facts with unseen timestamps.

\noindent \textbf{Textual Descriptions and LLMs}
SST-BERT \citep{chen2023incorporating}, ECOLA \citep{han2023ecola}, and PPT \citep{xu2023pre} utilize BERT-based architectures by pretraining on a TKG corpus, encoding additional textual information of facts, and converting link prediction into masked token prediction, respectively. ICL \citep{lee2023temporal} explores in-context learning of temporal facts within LLMs, while GenTKG \citep{liao2024gentkg} encodes temporal paths retrieved through logical rule-based methods in LLMs. zrLLM \citep{ding2024zrllm} first generates relation representations by inputting their textual descriptions into LLMs. All these methods demand substantial computational resources or rely on text descriptions of entities and relations. 
In contrast, our method is purely structure-driven: it learns directly from the topology of temporal knowledge graphs and does not require any textual annotations for entities or relations.


%% file: latex/sec_methodology.tex
\section{Methodology}

To enable fully inductive inference on TKGs, we propose the \ModelName{} model. The overall architecture of the model is illustrated in Figure~\ref{fig:model_overview}.

\subsection{Relational Structure Learning}
\label{sec:relation_learning}

To enable the relational transferability across datasets, we adopt the relation encoding method proposed by ULTRA \citep{galkintowards}. ULTRA constructs a relation-relation interaction graph for every knowledge graph with four common types of interactions between relations in the knowledge graph: $H=\{h2h, h2t, t2h, t2t\}$, which means head-to-head, head-to-tail, tail-to-head, and tail-to-tail interactions. \textit{Interactions of Relations Construction} part in Fig. \ref{fig:model_overview} illustrates these interactions. 



Given a TKG $G$, we construct a \emph{relation graph} $G_r=(R, H, O)$. 
In $G_r$ the nodes are the relation names of $G$, and the edges are the interactions of these relations. 
For example, $(p_1, h2h, p_2)$ belongs to $O$ if and only if there are two temporal facts of the form $(v_1,p_1,v_2, \tau_1)$ and $(v_1, p_2, v_3, \tau_2)$ in graph $G$.


Given a temporal query $(s, p, ?, \tau)$ and a relation graph ${G}_r$, we then obtain $d$-dimensional relation representations $\mathbf{r}_q, q \in R$ via message-passing with neighboring relation nodes:
\begin{align}
\mathbf{r}_{q|p}^0 &
= \mathbf{1}_{p=q} * \mathbf{1}^d, \quad q \in R  \\
\mathbf{r}_{q|p}^{l+1} &=  
\operatorname{AGG} \Big( 
\operatorname{MSG}(\mathbf{r}_{w|p}^l, \mathbf{h}) \mid  w \in {N}_h(p), h \in H
\Big) 
\label{eq:ultra-agg}
\end{align}
%
%
%
%
where $\mathbf{r}_{q|p}^{l+1}$ stands for the relation $q$ representation at $(l+1)$-th layer.
$N_{h}(p)$ stands for the neighboring relation nodes of $p$ which are connected by interations in $H$.
$\operatorname{AGG}$ is the aggregation function (e.g., $sum$), and $\operatorname{MSG}$ is the message function (e.g., DistMult's MSG \citep{yang2015embedding} is $\mathbf{r}_{w|p}^l * \mathbf{h}$, TransE's MSG \citep{bordes2013translating} is $\mathbf{r}_{w|p}^l + \mathbf{h}$), which defines how information is passed from a node's neighbors to that node.
$\mathbf{r}_{q|p}^{0}$ is initialized by detecting if the relation node is equal to the query relation $p$. The GNN architecture follows Neural Bellman-Ford Network\citep{zhu2021neural}. As these four interactions are universal and independent of datasets, the relational transferability can be achieved by transferring the embeddings of the four interactions.


\subsection{Temporal Embedding}
\label{sec:temporal_embedding}
While previous works have used universal structures in graphs as vocabularies for pretraining Graph Foundation Models (GFM) \citep{sun2025riemanngfm, wang2024gft}, temporal information differs fundamentally from nodes and edges in graphs, as time is inherently sequential. 
To address this, we represent a temporal knowledge graph (TKG) as a sequence of TKG snapshots $G_i$, where timestamp index $i$ denotes the position of $G_i$ in the sequence. 
Given that sinusoidal positional encodings have been shown to effectively capture sequential dependencies, we adopt a similar structure to encode temporal information in TKG sequences.
Specifically, we utilize sine and cosine functions with different frequencies to encode the temporal information for $G_i$ with the same dimension size as the proposed model.
\begin{align*}
     &[\operatorname{TE}(i)]_{2n}= \sin(\omega_n i),
    \qquad
    [\operatorname{TE}(i)]_{2n+1}= \cos(\omega_n i),\\
    & \omega_n \in R, \quad 0\le  n<\tfrac{d_{\operatorname{PE}}}{2},
\end{align*}
where $n$ is the dimension index of the embedding. 
$\omega_n$ can be trainable or set to 
$\omega_n = \beta^{-2n/d_{\operatorname{PE}}}$
where $d_{PE}$ is the dimension size of the temporal embedding and quadruple representation. 
 We also denote representation of $\tau_i$ by $\operatorname{TE}_{\tau_i} = \operatorname{TE}(i)$ where $i$ is the timestamp index.
The original Transformer architecture sets $\beta$ as 10000. In this way, each dimension of the positional encoding corresponds to a sinusoidal function, with wavelengths forming a geometric progression from $2\pi$ to $ 10000\cdot2\pi$. 
We chose this function for its computational simplicity and dataset‑agnostic design, which preserves temporal transferability.
Moreover, prior work has demonstrated that such positional encodings enable the model to effectively learn relative position dependencies \citep{su2024roformer,vaswani2017attention}, which we argue is crucial for capturing temporal dependencies between temporal facts. 

\subsection{Temporal-Aware Quadruple Learning}
\label{sec:entity_learning}
\parabold{Global Quadruple Representation}
\label{sec:global_graph}
We incorporate temporal information by integrating the temporal embedding into quadruple-level message passing. For a given temporal query $(s, p, ?, \tau_i)$, we first derive the $d$-dimensional entity representation $\mathbf{e}_{v|s}, v \in G$ on the entity graph conditioned on the temporal query with head entity $s$, following a similar approach as described in Section \ref{sec:relation_learning}: 
\begin{align}
\label{eq:tmsgglobal}
\mathbf{e}_{v|s}^0 &
= \mathbf{1}_{v=s} * \mathbf{r}_p, \quad v \in {G}, \\
\mathbf{e}_{v|s}^{l+1} &=  
\operatorname{AGG} \Big( 
\operatorname{T-MSG}(\mathbf{e}_{w}^l, \mathbf{r}_q, \sum{g^{l+1}(\operatorname{TE}_{\tau_j}})) \mid  w \in {N}_q(v), \, q \in R, \forall \tau_j : (e_w, q, v, \tau_j) \in G
\Big) ,
\end{align}
where $\mathbf{e}_{v|s}^{l+1}$ is the quadruple representation at the $(l+1)$th layer, $N_{p}(s)$ denotes the neighboring entity nodes of $s$, and $\operatorname{TE}_{\tau_j}$ embedding of all timestamp indices where $(s,p,v, \tau_j)$ occurs in the training graph during training and in the inference graph during testing. $g^{l+1}(.)$ is a linear function defined per layer.
$\operatorname{T-MSG}$ is a temporal aggregation mechanism that incorporates the temporal embedding described in Section \ref{sec:temporal_embedding}. In our experiments, we employ TTransE \citep{leblay2018deriving}, TComplEx \citep{tcomplexlacroix2020tensor}, and TNTComplEx\citep{tcomplexlacroix2020tensor} as temporal modeling approaches. The initial entity representation $e_{v|s}^{0}$ is determined by checking whether the entity node $v$ matches the query entity $s$. If so, it is initialized using the relation representation $\mathbf{r}_p$ learned in Section \ref{sec:relation_learning}. Since both the relation representations from Section \ref{sec:relation_learning} and the temporal representations from Section \ref{sec:temporal_embedding} are transferable, the learned entity embeddings also satisfy the transferability requirement.



\parabold{Local Quadruple Representation}
Relations in TKGs may exhibit different frequencies of change, ranging from fully static to rapidly evolving behaviors \citep{tcomplexlacroix2020tensor}. For example, the relation \textit{CapitalOf} tends to remain stable over time, whereas the relation \textit{Consult} is typically short-term. To effectively model these dynamic relations, which are more influenced by adjacent quadruples, we construct a local graph for a given temporal query $(s, p, ?, \tau_i) \in G_i$. 
Specifically, we define the local graph as $\Glocal = \{G_{i-k}, ..., G_{_i},..., G_{i+k}\} $, where $k$ is time window size.
The local quadruple representation $e_{v|s, \tau_i,local}$ is calculated as: 
\begin{equation}
\begin{split}
&\mathbf{e}_{v|s, \tau_i,local}^0 
= \mathbf{1}_{v=s} * \mathbf{r}_p, \quad v \in {G_{local}}, \\
&\mathbf{e}_{v|s, \tau_i,local}^{l+1} =  
\operatorname{AGG} \Big( 
\operatorname{T-MSG}(\mathbf{e}_{w,\text{local}}^l, \mathbf{r}_q, g^{l+1}(TE_{\tau_j}) \mid  w \in {N}_q(v), \, q \in R, \, \forall \tau_j \in T \, (e_w, q, v, \tau_j) \in G_{\text{local}}
\Big).
\end{split}
\end{equation}

Here, we reuse the message-passing structure from \autoref{eq:tmsgglobal} to minimize the number of parameters while maintaining consistency in the model architecture.

\parabold{Prediction}
Given a temporal query $(s, p, ?, \tau_i)$ and a tail candidate $o_{cand}$, the final score is:
\begin{equation}
\begin{split}
    &V_{s,p,o_{cand}} = \alpha \, \mathbf{e}_{o_{cand} |s, \tau_i,local} + (1-\alpha) \, \mathbf{e}_{o_{cand}|s}, \\
    &S(s, p, o_{cand},\tau_i) = f_\theta\bigl(V_{s,p,o_{cand}},\,\mathrm{TE}_{\tau_i}\bigr),
    \label{scorefunctionequation}
\end{split}
\end{equation}
where $V_{s,p,o_{cand}}$ is the time-aware triple representation (from message passing), $f_\theta$(.) is a multilayer perceptron with parameters $\theta$, and $\alpha$ is a hyperparameter, providing a tradeoff between local and global information. 
The method offers theoretical benefits, including the ability to model relations occurring at different temporal frequencies; formal theorems and their proofs are in Appendix \ref{sec:theoretical_analysis}.

%% file: latex/sec_experiment_setup.tex
\section{Experimental Setup}
\label{sec:experiment_setup}
\parabold{Datasets}
To evaluate the effectiveness of the proposed model on the fully-inductive inference task, we conduct link prediction experiments on five widely-used temporal knowledge graph benchmark datasets: ICEWS14~\citep{garcia-duran-etal-2018-learning}, ICEWS05-15~\citep{garcia-duran-etal-2018-learning}, GDELT~\citep{trivedi2017know}, ICEWS18~\citep{jin2020recurrent}, and YAGO~\citep{jin2020recurrent}. These datasets cover a diverse range of domains and temporal characteristics:

\begin{enumerate}
    \item \textbf{Inference Type:} ICEWS14, ICEWS05-15, and GDELT were designed for transductive inference, while ICEWS18 and YAGO were developed for temporal extrapolation tasks.
    
    \item \textbf{Domain Coverage:} ICEWS14, ICEWS05-15, and ICEWS18 are subsets of the Integrated Crisis Early Warning System (ICEWS)~\citep{lautenschlager2015icews}, consisting of news event data. GDELT is a large-scale graph focused on capturing human behavior events. YAGO is derived from the YAGO knowledge base for commonsense facts.
    
    \item \textbf{Temporal Granularity:} ICEWS14, ICEWS05-15, and ICEWS18 use daily timestamps. GDELT provides fine-grained 15-minute intervals, while YAGO uses yearly granularity.
    
    \item \textbf{Time Span:} ICEWS14 covers the year 2014, ICEWS05-15 spans from 2005 to 2015, and ICEWS18 includes events from January 1 to October 31, 2018. GDELT spans one year from April 2015 to March 2016, while YAGO spans a long range of 189 years.
\end{enumerate}

To assess the model’s fully-inductive inference capabilities, we perform cross-dataset zero-shot evaluations, where the model is trained on one dataset and evaluated on another, ensuring no overlap in entities, relations, or timestamps. Table \ref{table:dataset} in the Appendix shows the detailed splits of the datasets.



\parabold{Baselines}
We compare \ModelName{} with two SOTA fully-inductive inference models, INGRAM\citep{lee2023ingram} and ULTRA\citep{galkintowards}, which can handle unknown entities and relations for fully-inductive inference. We also compare with two LLM-based temporal knowledge graph reasoning models, ICL\citep{lee2023temporal} and GenTKG\citep{liao2024gentkg}, which focus on temporal extrapolation inference task. 




\parabold{Evaluation Metrics}
We adopt the link prediction task to evaluate our proposed model. Link prediction infers the missing entities for incomplete facts. 
During the test step, we follow the procedure of \citep{xu2020tero} to generate candidate quadruples. From a test quadruple $(s,p,o, \tau)$, we replace $s$ with all $\bar{s} \in V$ and $o$ with all $\bar{o} \in V$ to get candidate answer quadruples $(s,p,\bar{o}, \tau)$ and $ (\bar{s},p,o, \tau)$ to queries $(s,p,?,\tau)$ and $(?, p,o,\tau)$.  
For each query, all candidate answer quadruples will be ranked by their scores using a time-aware filtering strategy \citep{goel2020diachronic}. We evaluate our models with four metrics: Mean Reciprocal Rank (MRR), the mean of the reciprocals of predicted ranks of correct quadruples, and Hits@(1/10), the percentage of ranks not higher than 1/10. For all experiments, the higher the better. For ICEWS18 and YAGO, we perform the single-step prediction as mentioned in \citep{gastinger2023comparing}. We provide the hyperparameters and training details in Section \ref{sec:parameter} in the appendix.

%% file: latex/sec_result.tex
\begin{table*}[t!]
\centering
\caption{Fully-inductive link prediction results. Each block shows training on one dataset and testing on the remaining without fine-tuning. Best results are in \textbf{bold}.}
\label{tbl:LinkPredictionResults}
\begin{minipage}{\textwidth}
\begin{adjustbox}{width=\textwidth}
\begin{tabular}{lcccccccccccc|| ccc}
\toprule
\multicolumn{16}{c}{\textbf{Trained on ICEWS14}} \\
\midrule
\multirow{2}{*}{Model} 
& \multicolumn{3}{c}{\textbf{To ICEWS05-15}} 
& \multicolumn{3}{c}{\textbf{To GDELT}} 
& \multicolumn{3}{c}{\textbf{To ICEWS18}} 
& \multicolumn{3}{c}{\textbf{To YAGO}}
& \multicolumn{3}{c}{\textbf{Total Avg}}\\
& MRR & H@1 & H@10 & MRR & H@1 & H@10 & MRR & H@1 & H@10 & MRR & H@1 & H@10 & MRR & H@1 & H@10\\
\midrule
INGRAM      & 6.3  & 1.3  & 16.3 & 7.7  & 3.0  & 15.9 & 10.3 & 7.0  & 13.4 & 17.5 & 13.2 & 25.6 & 10.5 & 6.1 & 17.8 \\
ULTRA       & \underline{11.1} & \underline{12.0} & \underline{15.0} & \underline{17.0} & \underline{10.0} & \underline{30.0} & \underline{21.8} & \underline{11.5} & \underline{43.4} & \underline{63.7} & \underline{50.0} & \underline{91.5} & \underline{28.4} & \underline{21.0} & \underline{45.0}\\
\midrule
\textbf{\ModelName{}} & \textbf{42.9} & \textbf{35.3} & \textbf{55.3} & \textbf{26.1} & \textbf{17.2} & \textbf{43.8} & \textbf{25.0} & \textbf{14.2} & \textbf{47.3} & \textbf{87.9} & \textbf{83.9} & \textbf{93.2} & \textbf{45.5} & \textbf{37.7} & \textbf{59.9}\\
\bottomrule
\end{tabular}
\end{adjustbox}
\end{minipage}


\centering
\begin{minipage}{\textwidth}
\begin{adjustbox}{width=\textwidth}
\begin{tabular}{lcccccccccccc|| ccc}
\toprule
\multicolumn{16}{c}{\textbf{Trained on ICEWS05-15}} \\
\midrule
\multirow{2}{*}{Model} 
& \multicolumn{3}{c}{\textbf{To ICEWS14}} 
& \multicolumn{3}{c}{\textbf{To GDELT}} 
& \multicolumn{3}{c}{\textbf{To ICEWS18}} 
& \multicolumn{3}{c}{\textbf{To YAGO}}
& \multicolumn{3}{c}{\textbf{Total Avg}}\\
& MRR & H@1 & H@10 & MRR & H@1 & H@10 & MRR & H@1 & H@10 & MRR & H@1 & H@10 & MRR & H@1 & H@10 \\
\midrule
INGRAM      & 11.1 & 3.8  & 27.8 & 3.8  & 1.0  & 7.1  & 6.2  & 2.0  & 15.3 & 12.4 & 7.6  & 21.2 & 8.4 & 3.6 & 17.9 \\
ULTRA       & \underline{46.5} & \underline{34.3} & \underline{70.4} & \underline{17.9} & \textbf{10.2} & \underline{32.4} & \underline{14.5} & \underline{8.3}  & \textbf{36.1} & \underline{61.2} & \underline{46.6} & \textbf{90.9} & \underline{35.0} & \underline{24.9} & \textbf{57.5}\\
\midrule
\textbf{\ModelName{}} & \textbf{51.5} & \textbf{41.5} & \textbf{70.6} & \textbf{18.9} & \textbf{10.2} & \textbf{36.7} & \textbf{15.8} & \textbf{8.7}  & \underline{29.7} & \textbf{62.2} & \textbf{49.0} & \underline{86.5} 
& \textbf{37.1} & \textbf{27.4} & \underline{55.9}\\
\bottomrule
\end{tabular}
\end{adjustbox}
\end{minipage}


\centering
\begin{minipage}{\textwidth}
\begin{adjustbox}{width=\textwidth}
\begin{tabular}{lcccccccccccc|| ccc}
\toprule
\multicolumn{16}{c}{\textbf{Trained on GDELT}} \\
\midrule
\multirow{2}{*}{Model} 
& \multicolumn{3}{c}{\textbf{To ICEWS14}} 
& \multicolumn{3}{c}{\textbf{To ICEWS05-15}} 
& \multicolumn{3}{c}{\textbf{To ICEWS18}} 
& \multicolumn{3}{c}{\textbf{To YAGO}} 
& \multicolumn{3}{c}{\textbf{Total Avg}}\\
& MRR & H@1 & H@10 & MRR & H@1 & H@10 & MRR & H@1 & H@10 & MRR & H@1 & H@10 & MRR & H@1 & H@10 \\
\midrule
INGRAM      & 8.6  & 2.9  & 20.7 & 5.8  & 1.2  & 14.4 & 5.0  & 2.3  & 9.3  & 7.4  & 3.3  & 15.0 & 6.7 & 2.4 & 14.9 \\
ULTRA       & \underline{33.6} & \underline{24.5} & \underline{50.7} & \underline{38.3} & \underline{26.8} & \underline{60.5} & \underline{24.4} & \underline{15.6} & \underline{41.8} & \underline{48.5} & \underline{40.1} & \underline{63.2} & \underline{36.2} & \underline{26.8} & \underline{54.1}\\
\midrule
\textbf{\ModelName{}} & \textbf{43.7} & \textbf{32.4} & \textbf{65.7} & \textbf{45.5} & \textbf{33.1} & \textbf{70.0} & \textbf{27.2} & \textbf{16.2} & \textbf{50.4} & \textbf{63.3} & \textbf{50.2} & \textbf{86.8} & \textbf{44.9} & \textbf{33.0} & \textbf{68.0}\\
\bottomrule
\end{tabular}
\end{adjustbox}
\end{minipage}
\end{table*}

\section{Experimental Result}
\label{sec:result}
\subsection{Fully-Inductive Inference Results}
We evaluate \ModelName{}’s fully-inductive inference performance across multiple zero-shot scenarios, where no fine-tuning is applied during testing, as presented in Table \ref{tbl:LinkPredictionResults}. \ModelName{} consistently outperforms baseline models across multiple evaluation metrics, demonstrating its strong fully-inductive learning capability. From the result, we have following observations: 

1) \ModelName{} generalizes well to datasets with varying time spans and temporal granularities. The time span of the experimental datasets ranges from just 1 month (GDELT) to 189 years (YAGO), while their temporal granularities vary from 15 minutes (GDELT) to 1 year (YAGO). Despite being trained on one dataset and evaluated on another with significantly different temporal characteristics, \ModelName{} consistently achieves strong performance across all scenarios. This demonstrates the robustness and generalization ability of our sequential temporal embedding in handling diverse and unseen temporal information. Furthermore, the performance gap between \ModelName{} and ULTRA is more pronounced when trained on ICEWS14 compared to ICEWS05-15, suggesting that the proposed sequential temporal embedding is particularly beneficial for smaller datasets.

2) \ModelName{} generalizes well across datasets from different domains and varying densities. The experimental datasets cover a wide range of domains, from encyclopedic knowledge in YAGO to diplomatic event data in ICEWS. \ModelName{} achieves impressive results even when trained on one domain and evaluated on another (see ``Trained on ICEWS14 to YAGO''), demonstrating the model’s ability to transfer across domains. This highlights that the learned quadruple representations from the relation encoder and quadruple encoder are capable of capturing domain-specific structural knowledge. Moreover, the datasets also vary in density: GDELT has more frequent events per timestamp, while ICEWS14 is relatively sparse. \ModelName{} consistently performs well in cross-dataset evaluation, demonstrating its robustness in event densities.

3) \ModelName{} generalizes well to both temporal knowledge graph interpolation and extrapolation tasks. As shown in Table \ref{tbl:Comparison_with_LLM}, \ModelName{} achieves strong results when trained to predict historical unseen events and tested on future unseen events, even outperforming LLM-based models, which are designed for extrapolation tasks. Notably, the cost of LLM-based models to compute the standard MRR metric over all candidate entities is prohibitive, whereas \ModelName{} efficiently supports full evaluation. This highlights both the flexibility and the computational efficiency of \ModelName{}.


\begin{table}[H]
    \centering
    \caption{
        Link prediction results on ICEWS18 and YAGO between LLM-based models and \ModelName{} trained on ICEWS14.
        The best results among all models are in \textbf{bold}.
    }
    \label{tbl:Comparison_with_LLM}
    \begin{tabular}{lcccccc}
        \toprule
        \multirow{2}{*}{\textbf{Model}} 
        & \multicolumn{3}{c}{\textbf{ICEWS18}} 
        & \multicolumn{3}{c}{\textbf{YAGO}} \\
        \cmidrule(lr){2-4} \cmidrule(lr){5-7}
        & MRR & H@1 & H@10 & MRR & H@1 & H@10 \\
        \midrule
        ICL         & --         & 18.2       & 41.4       & --         & 78.4       & 92.7 \\
        GenTKG      & --         & \textbf{24.3} & 42.1    & --         & 79.2       & 84.3 \\
        \midrule
        \ModelName{}  & \textbf{25.0} & 14.2   & \textbf{47.3} & \textbf{87.9} & \textbf{83.9} & \textbf{93.2} \\
        \bottomrule
    \end{tabular}
\end{table}

\subsection{Ablation Study}

\begin{wraptable}{r}{0.5\textwidth}
    \centering
    \caption{
        Ablation study of \ModelName{}. 
        TE denotes Temporal Encoding, GQR denotes Global Quadruple Encoder, and LQR denotes Local Quadruple Encoder. 
        All models are trained on ICEWS14 and evaluated on ICEWS05-15.
    }
    \label{tbl:Ablation_Study}
    \begin{tabular}{lccc}
        \toprule
        \textbf{Model} & \textbf{MRR} & \textbf{H@1} & \textbf{H@10} \\
        \midrule
        \ModelName{}      & \textbf{42.9} & \textbf{35.3} & 55.3 \\
        \midrule
        w/o TE          & 11.1 & 12.0 & 15.0 \\
        w/o GQR         & 40.0 & 32.5 & 52.6 \\
        w/o LQR         & 37.9 & 25.6 & \textbf{62.6} \\
        \bottomrule
    \end{tabular}
\end{wraptable}

\paragraph{Module Ablation} To validate the effectiveness of each module in \ModelName{}, we conducted ablation studies on three key components: Temporal Embedding (TE), Global Quadruple Representation (GQR), and Local Quadruple Representation (LQR). As shown in Table \ref{tbl:Ablation_Study}, the results highlight the importance of each component. Removing the Temporal Embedding module results in a substantial performance drop, emphasizing Temporal Embedding's transferability between different temporal knowledge graphs. Excluding the Global Quadruple Representation significantly reduces the model's Hits@10 score, supporting our hypothesis that it captures long-term global temporal information. Similarly, removing the Local Quadruple Representation notably decreases the Hits@1 score, aligning with our assumption that it models the concurrent interactions of temporal events. We provide a more detailed visualization of GQR and LQR in Section \ref{sec:case_study} in the Appendix.



\parabold{Performance on Temporal Structural Patterns} To further assess the model’s ability to capture temporal structural patterns, we evaluate on two temporal structural patterns: symmetric and inverse. We construct subsets from the ICEWS05-15 test set that contain such structural patterns\footnote{Symmetric and inverse subsets contain 12,600 and 28,891 quadruples, respectively. Data splits are provided in our codebase under dataset classes \texttt{ICEWS0515\_sym} and \texttt{ICEWS0515\_inv}.} and report results in Table~\ref{tbl:Pattern_performance}. A pair of quadruples is \textit{symmetric} if for $(s, p, o, \tau_1)$, $(o, p, s, \tau_2)$ exists. A relation pair $(p, p')$ is \textit{inverse} if $(s, p, o, \tau_1)$ implies $(o, p', s, \tau_2)$. For both patterns, \ModelName{} consistently outperforms the baseline ULTRA. These results demonstrate that incorporating relative temporal signals significantly enhances the model's ability to learn and generalize structural patterns in TKGs.

\begin{table}[H]
    \centering
    \caption{
   Model performance on structural pattern subsets. All models are trained on ICEWS14. 
    }
    \label{tbl:Pattern_performance}
    \begin{tabular}{lcccccc}
        \toprule
        \multirow{2}{*}{\textbf{Model}} 
        & \multicolumn{3}{c}{\textbf{symmetric}} 
        & \multicolumn{3}{c}{\textbf{inverse}} \\
        \cmidrule(lr){2-4} \cmidrule(lr){5-7}
        & MRR & H@1 & H@10 & MRR & H@1 & H@10 \\
        \midrule
        ULTRA      & 53.5         & 38.8 & 81.8    & 45.7         & 32.5       & 70.9 \\
        \ModelName{}  & \textbf{67.0} & \textbf{53.7}   & \textbf{90.7} & \textbf{61.8} & \textbf{48.9} & \textbf{86.0} \\
        \bottomrule
    \end{tabular}
\end{table}

%% file: latex/sec_appendix.tex
\section{Parameter Count and Complexity}
From Table~\ref{table:time and space}, we observe that the number of parameters in \ModelName{} is agnostic to the dataset size. Since \ModelName{} does not initialize embeddings based on $|V|$, $|R|$, or $|\Tau|$, but instead relies on a fixed number of relation interactions $|H|$, its parameter count is only related to the embedding dimension $d$. This property makes \ModelName{} especially suitable for transfer learning scenarios.

The time complexity of \ModelName{} is primarily determined by the quadruple encoder, as the number of relations $|R|$ is significantly smaller than the number of entities $|V|$, allowing us to omit the complexity contribution of the relation encoder. Furthermore, since the local entity graph is much smaller than the global entity graph, the computational upper bound is dominated by the global quadruple representation. Utilizing NBFNet~\citep{zhu2021neural} as the encoder, the time complexity for a single layer is $O(|Q|d + |V|d^2)$. For $A$ layers, the total time complexity becomes $O(A(|Q|d + |V|d^2))$.

The memory complexity of \ModelName{} is also linear in the number of edges, expressed as $O(A|Q|d)$, as the quadruple encoder maintains and updates representations for each edge in the temporal knowledge graph.

\begin{table*}[h]
\centering
    \caption{
    Parameter number and average training time for \ModelName{}.
    }
    \label{table:time and space}
{ 
\begin{tabular}{lccccc}
    \toprule
 Model & Datasets & $d$ & Parameter number\ \ & Batch Size &\ \ Average epoch time(h) \cr 
\midrule
 \multirow{3}{*}{\ModelName{}}  
 & ICEWS14&64& 247,937 & 16&1.8  \cr
 &ICEWS05-15& 64 &247,937& 2 &21.5 \cr
 &GDELT & 64 & 247,937 & 1 & 123.3 \cr
\bottomrule
\end{tabular}
}
\end{table*}

\begin{table}[H]
\centering
    \caption{
    Detailed statistics of datasets.
    }
    \label{table:dataset}
\begin{tabular}{lccccc}
  \toprule
 Dataset &ICEWS14&ICEWS05-15&GDELT&ICEWS18 & YAGO \cr
  \midrule
 Entities  &7,128 &10,488  &500 &23,033 & 10,623 \cr
Relations & 230 & 251  & 20& 256 & 10  \cr
Times & 365 & 4017 &366 & 303 &189 \cr
Train& 72,826 &386,962 & 2,735,685& 373,018 & 161,540 \cr
Validation & 8,941& 46,275 & 341,961& 45,995& 19,523 \cr 
Test & 8,963 & 46,092 & 341,961 & 49,995 & 20,026 \cr 
    Granularity & Daily & Daily & 15 minutes & Daily & Yearly  \cr
        \bottomrule
\end{tabular}
\end{table}

\section{Pre-training Results}
 Table~\ref{tbl:pretraining_results} presents the pretrained performance of \ModelName{}. It achieves comparable performance with TComplEx while using significantly fewer parameters especially on datasets which involve a large number of entities and relations. \ModelName{} generates adaptive representations dynamically through a conditional message passing mechanism. This design enables effective generalization with a compact model size, making it highly scalable and suitable for large temporal knowledge graphs.


\begin{table}[H]
    \centering
    \caption{ 
Pretrained results of POSTRA.
    }
    \label{tbl:pretraining_results}
    \resizebox{\textwidth}{!}{
    \begin{tabular}{lccccccccc}
        \toprule
        \multirow{2}{*}{\textbf{Model}} &
         \multicolumn{3}{c}{\textbf{ICEWS14}} &
         \multicolumn{3}{c}{\textbf{ICEWS-0515}} & \multicolumn{3}{c}{\textbf{GDELT}} \\
        \cmidrule(lr){2-4} \cmidrule(lr){5-7} \cmidrule(lr){8-10} 
        & MRR & H@10 & Parameter Count & MRR & H@10  & Parameter Count& MRR & H@10  & Parameter Count\\
        \midrule
        TComplEx         & 56.2 & 73.2 &2,035,968 &58.0 & 78.1 & 3,841,792 &21.4 &40.1 & 251,936\\
        \ModelName{}  & \textbf{58.3} & \textbf{76.0}& 247,937 & \textbf{59.0} & \textbf{78.7}&274,937 &\textbf{24.5} &\textbf{42.7} & 247,937\\
        \bottomrule
    \end{tabular}}
\end{table}


\section{Limitations}
\label{sec:limitation}
Although \ModelName{} demonstrates strong fully-inductive inference performance in zero-shot scenarios, several limitations remain. First, training on larger graphs does not always lead to improved performance. We hypothesize that variations in temporal patterns across datasets of different sizes may influence results. Second, \ModelName{} can be computationally intensive for dense temporal knowledge graphs, as its space complexity scales with the number of events. Third, \ModelName{} is unable to perform time prediction. We aim to explore more efficient architectures and adaptive training strategies to address these challenges in future work.

\section{Ethics Statement}
\label{sec:ethics_statement}
\ModelName{} can be applied to a wide range of tasks and datasets beyond their originally intended scope. Given the sensitive and time-dependent nature of temporal knowledge graphs (TKGs), there is a risk that such models could be used to infer patterns from historical or anonymized temporal data in ways that may compromise privacy or be exploited for surveillance, misinformation, or manipulation. On the positive side, fully-inductive models for TKGs offer significant benefits by enabling zero-shot transfer across domains, datasets, and temporal granularities. This reduces the need to train separate models for each new scenario, lowering the computational cost for training repeated models. 

\section{Relative Representation Transfer}
\label{sec:relative_representation}
\begin{figure}[htbp]
    \centering
    \begin{subfigure}[b]{0.32\textwidth}
        \includegraphics[width=\linewidth]{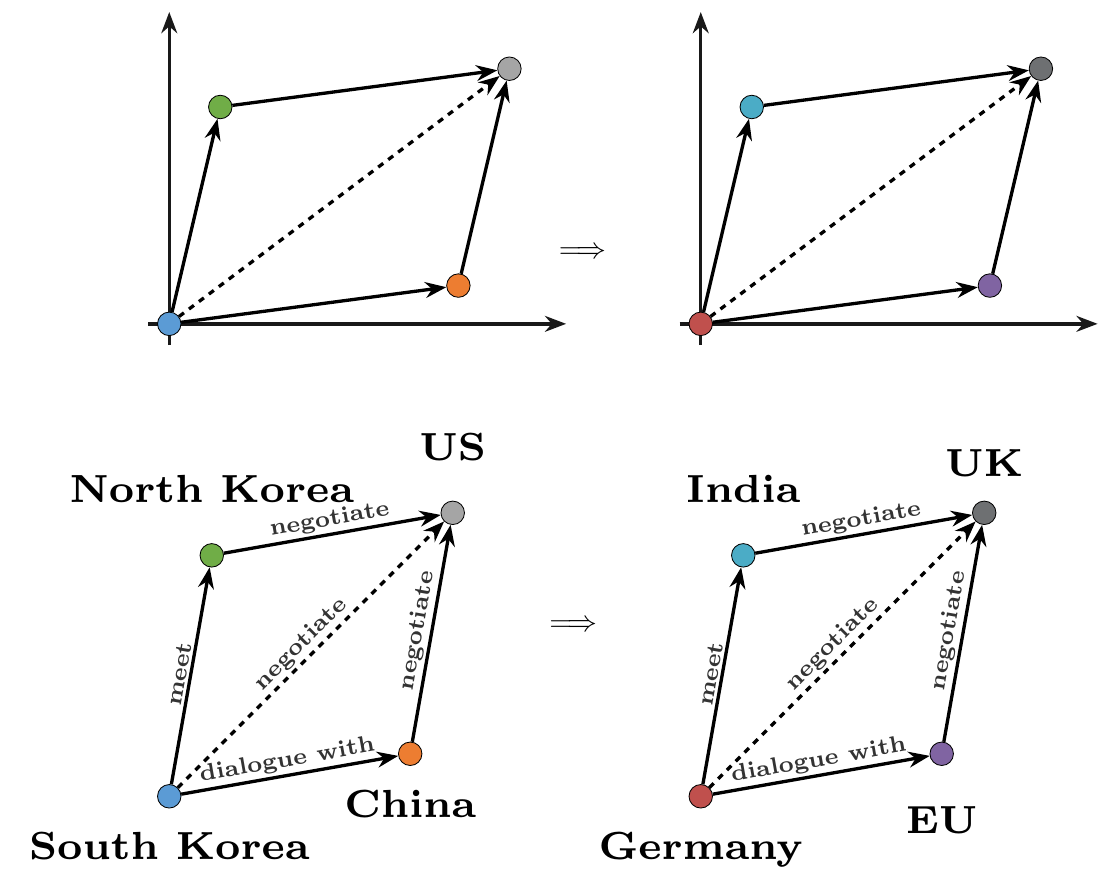}
        \caption{}
    \end{subfigure}
    \hfill
    \begin{subfigure}[b]{0.32\textwidth}
        \includegraphics[width=\linewidth]{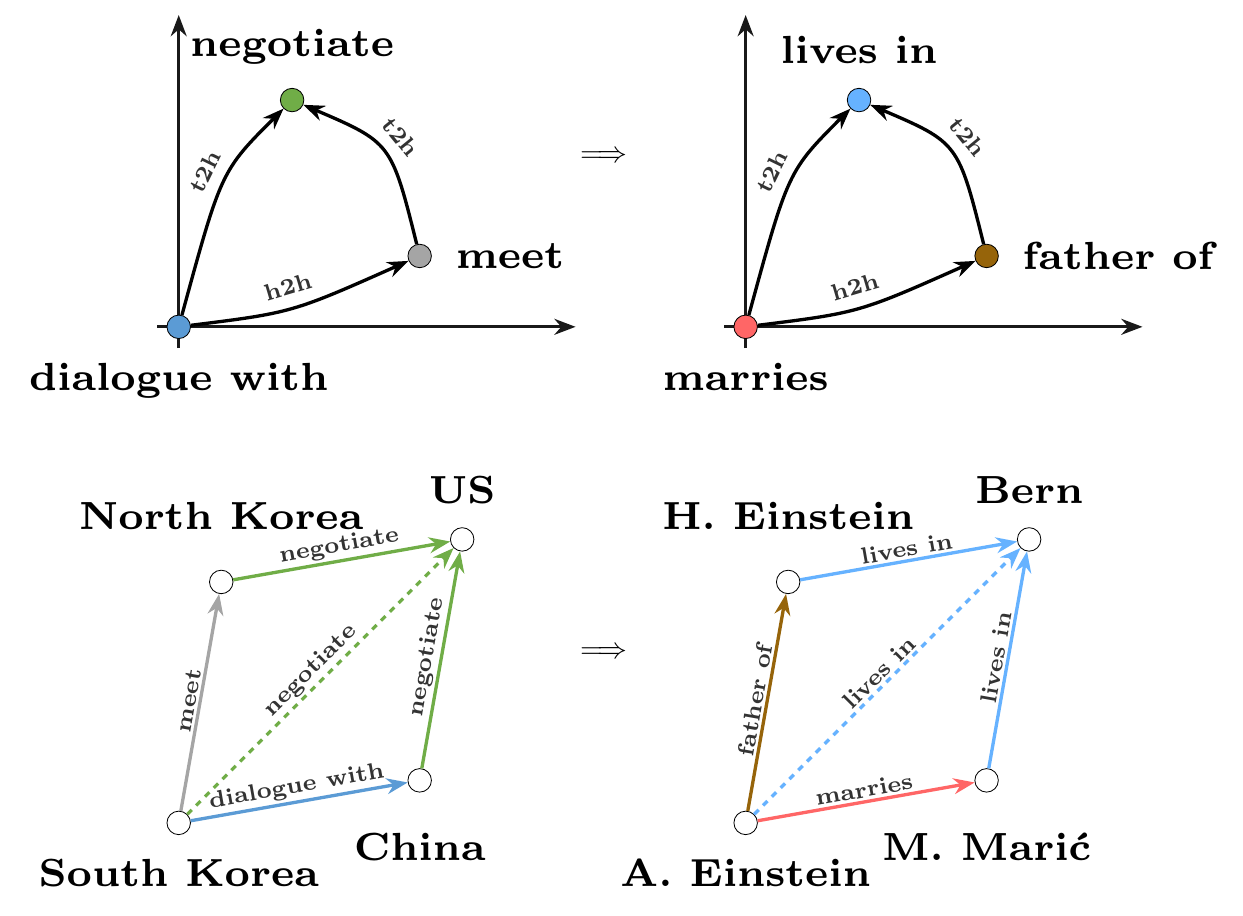}
        \caption{}
    \end{subfigure}
    \hfill
    \begin{subfigure}[b]{0.32\textwidth}
        \includegraphics[width=\linewidth]{figs/temporal_relation_transfer.pdf}
        \caption{}
    \end{subfigure}
    \caption{Relative Representation Transfer}
    \label{fig:relative_representation}
\end{figure}

 Subgraph (a) in Figure \ref{fig:relative_representation} shows the transfer of relative entity representations. In the left bottom KG, $(\text{South Korea} \xrightarrow{\text { meet }} \text{North Korea} \xrightarrow{\text { negotiate }} \text{US}) \wedge (\text{South Korea} \xrightarrow{\text { dialogue with }} \text{China} \xrightarrow{\text { negotiate }} \text{US})$ implies that $\text{South Korea} \xrightarrow{\text { negotiate }} \text{US}$. Given entity $v_1$, $v_2$, $v_3$, $v_4$, we can learn a structure that $(v_1 \xrightarrow{\text { meet }} v_2 \xrightarrow{\text { negotiate }} v_3) \wedge (v_1 \xrightarrow{\text { dialogue with }} v_4 \xrightarrow{\text { negotiate }} v_3)$ implies that $v_1 \xrightarrow{\text { negotiate }} v_3$ during training. In the right part, we substitute $v_1$, $v_2$, $v_3$, $v_4$ with Germany, India, UK and EU respectively, then we can infer that $\text{Germany} \xrightarrow{\text { negotiate }} \text{UK}$.
 
Subgraph (b) depicts the transfer of relative relation representations. In the left part, the upper relation graph depicts the interaction of relations in the bottom KG: $(\text{dialogue with} \xrightarrow{\text { h2h }} \text{meet} \xrightarrow{\text { t2h }} \text{negotiate}) \wedge (\text{dialogue with} \xrightarrow{\text { t2h }} \text{negotiate})$. Given relation $r_1$, $r_2$, $r_3$, we learn a relational structure: $(r_1 \xrightarrow{\text { h2h }} r_2 \xrightarrow{\text { t2h }} r_3) \wedge (r_1 \xrightarrow{\text { t2h }} r_3)$ during training. In the right part, we substitute $r_1$, $r_2$, $r_3$ with \textit{marries}, \textit{father of} and \textit{lives in} respectively, then we can the relational structure for the KG in the right part.

Subgraph (c) depicts the transfer of relative temporal ordering. In the left part, the upper relation graph depicts the relative temporal ordering of quadruples in the bottom TKG: $(\text{dialogue with} \xrightarrow{\Delta \text{=6} } \text{meet} \xrightarrow{\Delta \text{=4}} \text{negotiate}) \wedge (\text{dialogue with} \xrightarrow{\Delta \text{=5}} \text{negotiate})$. Given relation $r_1$, $r_2$, $r_3$, we learn a relational structure: $(r_1 \xrightarrow{\Delta \text{=6}} r_2 \xrightarrow{\Delta \text{=4}} r_3) \wedge (r_1 \xrightarrow{\Delta \text{=5}} r_3)$ during training. In the right part, we substitute $r_1$, $r_2$, $r_3$ with \textit{marries}, \textit{father of} and \textit{lives in} respectively, then we can the temporal transition information for the TKG in the right part.

\section{Hyperparameter and Training Details}
\label{sec:parameter}
Our experiments were conducted on 4 NVIDIA A100 GPUs with 48GB of RAM. We set the maximal training epoch as 10 and negative samples as 512. We use a batch size of 16, 2, and 1 for training ICEWS14, ICEWS05-15 and GDELT respectively. We set the dimension size $d$ as 64. We employed the AdamW optimizer and set the learning rate as 0.0005. More details can be seen in Table \ref{table:parameter}. For the baseline methods, we utilized the settings in their original papers. 

\begin{table}[t!]
\centering
    \caption{
    Detailed hyperparameters. $GNN_r$ denotes the relation representation encoder, $GNN_q$ is quadruple representation encoder.
    }
    \label{table:parameter}
{ 
\begin{tabular}{lccccc}
\toprule
Module & Hyperparameter & Pre-training  \cr
\hline
    \multirow{4}{*}{$GNN_r$} & \# layers & 6 \cr
   &  hidden dim & 64 \cr
   & MSG & Dual \cr
   & AGG & sum \cr
   \hline
    \multirow{5}{*}{$GNN_e$} & \# layers & 6 \cr
   &  hidden dim & 64 \cr
   & T-MSG & TComplEx \cr
   & AGG & sum \cr
   & $k$ & ICEWS14:0; ICEWS0515, GDELT:1 \cr
   & $\beta$ & 10,000 \cr
\hline   
  \multirow{2}{*}{Prediction} & $f_{\theta}$ & 2-layer MLP \cr
  & $\alpha$ & ICEWS14:0.5; ICEWS0515, GDELT:0.8 \cr
\bottomrule
\end{tabular}
}
\end{table}

\section{Hyperparameter Analysis}
\begin{figure}[htbp]
    \centering
    \begin{subfigure}[b]{0.32\textwidth}
        \includegraphics[width=\linewidth]{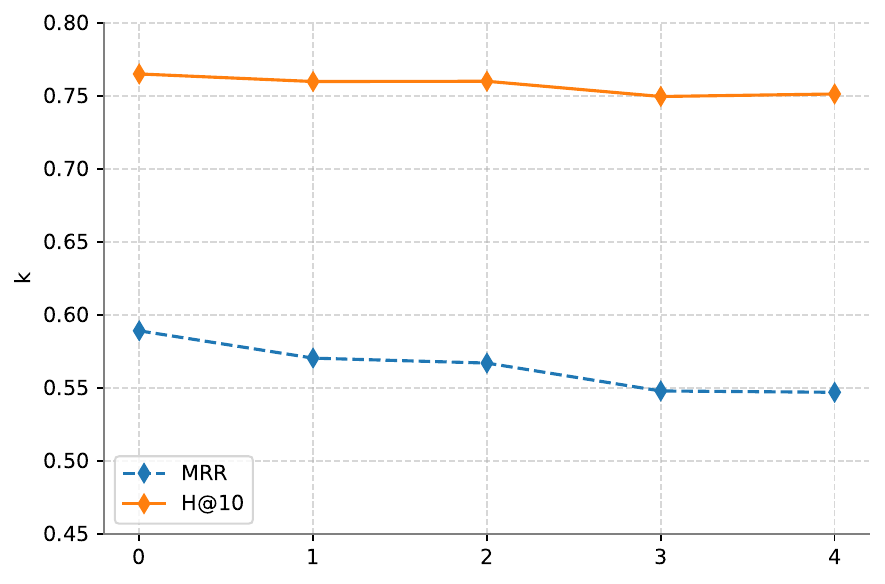}
        \caption{$k$'s performance}
    \end{subfigure}
    \hfill
    \begin{subfigure}[b]{0.32\textwidth}
        \includegraphics[width=\linewidth]{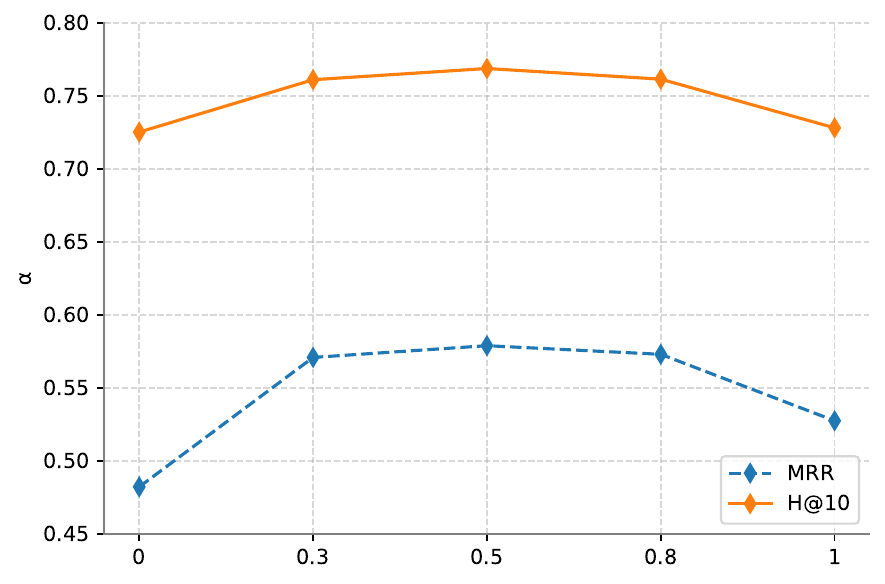}
        \caption{$\alpha$'s performance}
    \end{subfigure}
    \hfill
    \begin{subfigure}[b]{0.32\textwidth}
        \includegraphics[width=\linewidth]{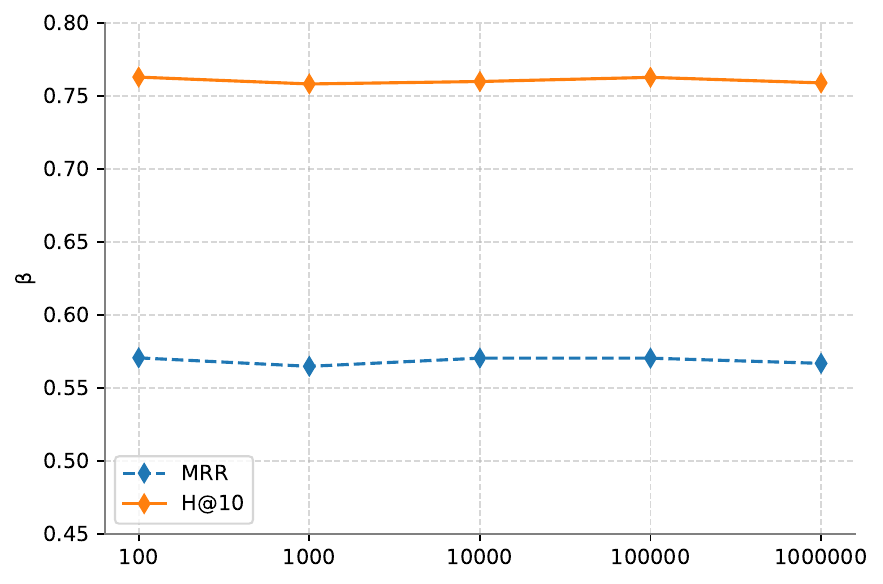}
        \caption{$\beta$'s performance}
    \end{subfigure}
    \caption{MRR and H@10 performance with different $k$, $\alpha$ and $\beta$. We report ICEWS14's pre-training performance and keep all other hyper-parameters as the fixed best setting.}
    \label{fig:hyperparamter_study}
\end{figure}

We conduct a sensitivity analysis of three key hyperparameters, $k$, $\alpha$, and $\beta$ on model performance by pre-training on the ICEWS14 dataset. Figure \ref{fig:hyperparamter_study} presents the MRR and H@10 results under different values of each hyperparameter while keeping all others fixed at their optimal settings.

\textbf{Effect of $k$.}  
As shown in Figure \ref{fig:hyperparamter_study}(a), increasing $k$, which controls the length of the local entity graph, leads to a gradual decline in both MRR and H@10. As ICEWS14 is a news dataset and contains mainly short-term temporal relations, deeper propagation may  cause oversmoothing and decrease the model's performance.

\textbf{Effect of $\alpha$.}  
Figure~\ref{fig:hyperparamter_study}(b) shows that performance is highly sensitive to the choice of $\alpha$, which balances the contribution of local and global temporal contexts. Both MRR and H@10 significantly improve as $\alpha$ increases from 0 to 0.5, indicating the benefit of incorporating both types of temporal information. However, performance degrades when $\alpha$ approaches 1, implying that overemphasis on either context harms generalization. The results highlight the importance of maintaining a moderate balance between local and global temporal patterns.

\textbf{Effect of $\beta$.}  
As illustrated in Figure \ref{fig:hyperparamter_study}(c), the model is highly robust to the choice of $\beta$, which controls the sinusoidal frequency of temporal embeddings. Across a wide range of values—from $10^2$ to $10^6$—MRR and H@10 remain largely stable, with negligible performance fluctuations. 

 
The hyperparameter analysis reveals that $\alpha$ is the most critical parameter and requires careful tuning to balance local and global temporal information. The choice of $k$ should be dataset-dependent; for datasets characterized by short-term temporal patterns, smaller values of $k$ are preferable. In contrast, the model is relative insensitive to $\beta$, so we set it as 10,000 across all datasets.

\section{Case Study}
\label{sec:case_study}
\begin{figure}
    \centering
    \begin{subcaptionbox}{Global Quadruple Representation \label{fig:first}}[0.48\textwidth]
        {\includegraphics[width=\linewidth]{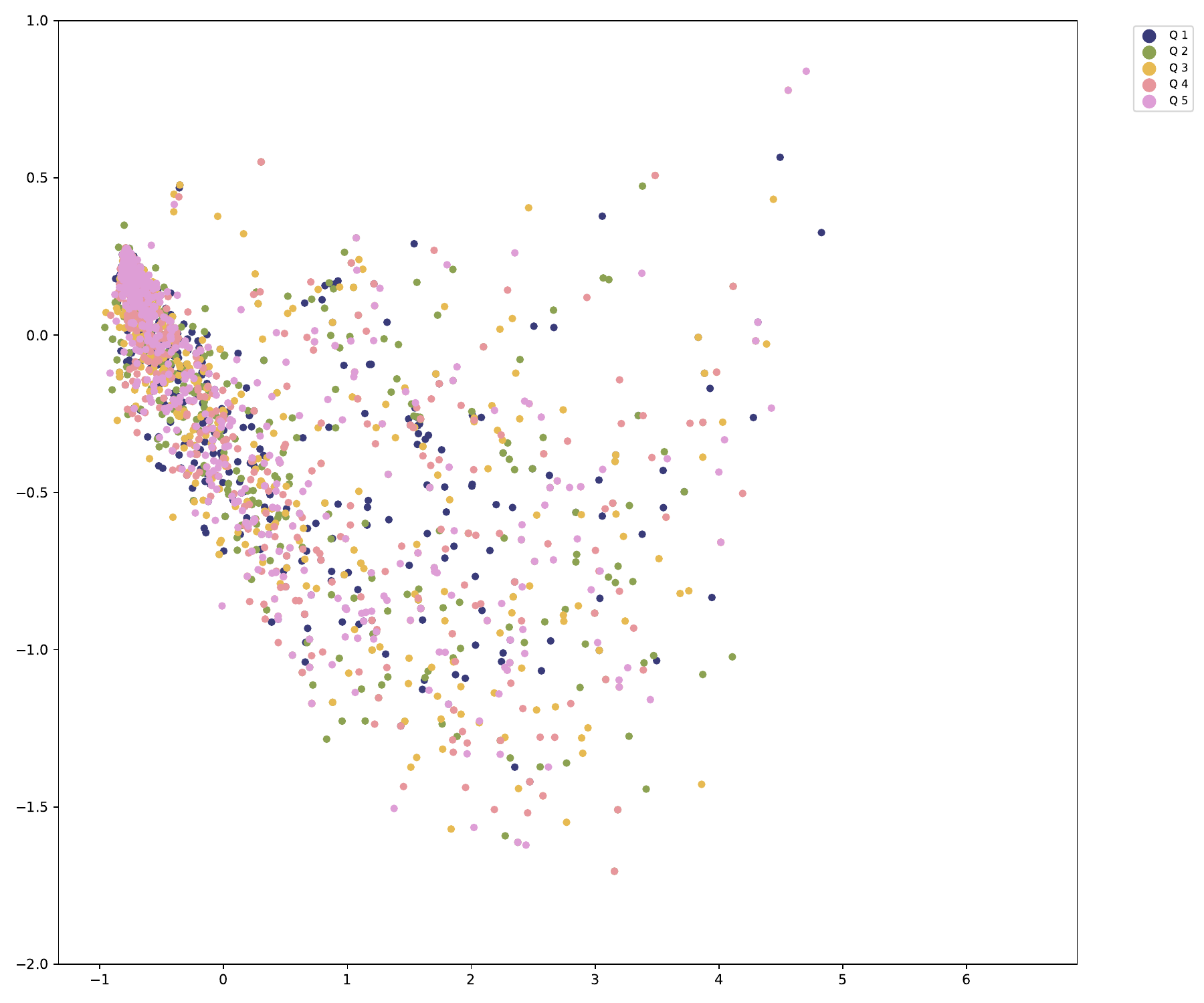}}
    \end{subcaptionbox}
    \hfill
    \begin{subcaptionbox}{Local Quadruple Representation\label{fig:second}}[0.48\textwidth]
        {\includegraphics[width=\linewidth]{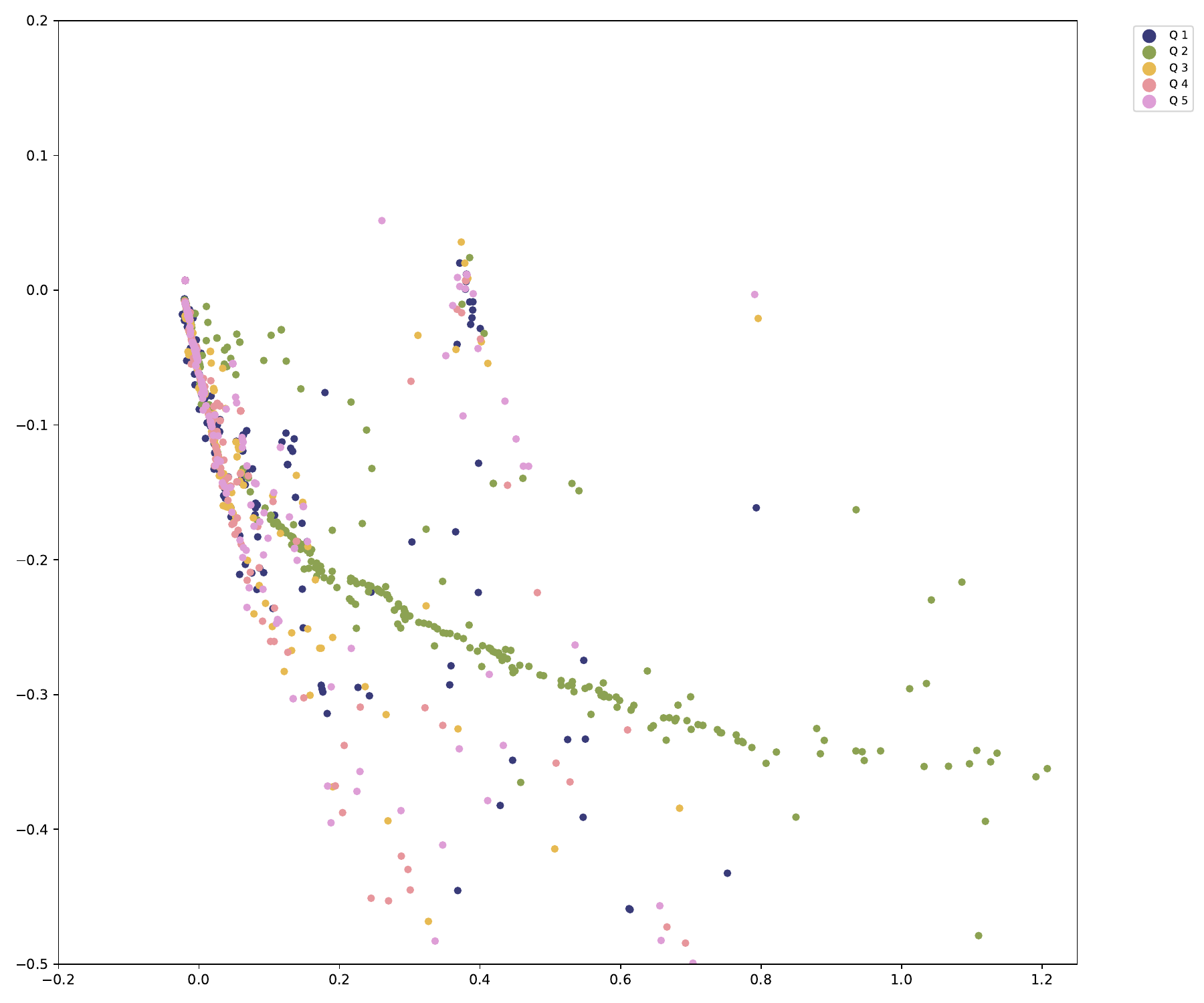}}
    \end{subcaptionbox}
    \caption{PCA visualizations of five quadruples with (China, Host a visit, ?, ?)}
    \label{fig:case_study}
\end{figure}

To intuitively demonstrate \ModelName{}’s capability in generating compact and distinguishable embeddings, we employ PCA to visualize the distribution of embeddings for five representative quadruples: Q1: (China, Host a Visit, Macky Sall, 2014-02-19), Q2:(China, Host a Visit, Teo Chee Hean, 2014-10-28), Q3: (China, Host a Visit, Chuck Hagel, 2014-04-08), Q4: (China, Host a Visit, Sar Kheng, 2014-06-03), and Q5: (China, Host a Visit, Barack Obama, 2014-12-25), as illustrated in Figure \ref{fig:case_study}. The visualization reveals that the Global Quadruple Representation tends to capture general relational patterns, producing embeddings that are closely clustered for the same query structure (China, Host a Visit, ?, ?). In contrast, the Local Quadruple Representation is more sensitive to immediate temporal context, resulting in more distinctive embeddings across different time points. This dual representation mechanism highlights \ModelName{}’s strength in capturing both global structural semantics and local temporal dynamics. For example, for Q5, Barack Obama is frequently the tail entity for similar queries, leading the global encoder to assign him a high probability and correctly predict the answer. Meanwhile, in Q1, a supporting local event (Macky Sall, Make a Visit, China, 2014-02-18) appears in the surrounding context, enabling the local quadruple encoder to identify Macky Sall correctly. However, the absence of any relevant local events involving Teo Chee Hean in the Q2 causes the model to mispredict Barack Obama.

\section{Theoretical Analysis}
\label{sec:theoretical_analysis}
In this section, we provide a theoretical analysis demonstrating the model’s ability to transfer across time and effectively capture a variety of periodic patterns.

\paragraph{Temporal Transferability}
As illustrated in \autoref{fig:temporal_relation_transfer} and \ref{fig:relative_representation}, the temporal difference is the transferable information in TKGs. Therefore, we expect that such transferability must be followed in the time embedding space. 
Here, we aim to show that if four time points' indices—$\tau_1, \tau_2$ in the training graph and $\tau'_1, \tau'_2$ in the test graph—hold $\Delta T = \tau_2 - \tau_1 = \Delta T' =  \tau'_2 - \tau'_1$, then we expect that such transferability is preserved via sinusoidal positional encoding in the form of Euclidean distance
\(\|\mathrm{TE}(\tau_2)-\mathrm{TE}(\tau_1)\| = \|\mathrm{TE}(\tau'_2)-\mathrm{TE}(\tau'_1)\|\)
that is governed solely by their time difference $\Delta T$, rather than the
absolute times.  

The following theorem states and proves this: 

\begin{theorem}[Time-Shift Invariance in Sinusoidal Positional Embeddings]
Let \(d\in\mathbb{N}\) be \emph{even} and fix positive frequencies
\(\omega_0,\omega_1,\dots,\omega_{\frac{d}{2}-1}>0\).
Define the \emph{sinusoidal temporal embedding}
\(\mathrm{TE}:\mathbb{N}\to\mathbb{R}^{d}\) component-wise by
\[
  \bigl[\mathrm{TE}(t)\bigr]_{2n}   \;=\;\sin\!\bigl(\omega_n\,t\bigr),
  \qquad
  \bigl[\mathrm{TE}(t)\bigr]_{2n+1} \;=\;\cos\!\bigl(\omega_n\,t\bigr),
  \quad 0\le n<\tfrac{d}{2}.
\]

For any four time-points indices \(\tau_1,\tau_2,\tau'_1,\tau'_2\in\mathbb{N}\) set
\[
  \Delta_\tau \;:=\; \tau_2-\tau_1,
  \qquad
  \Delta_\tau' \;:=\; \tau'_2-\tau'_1.
\]
If \(\Delta_\tau=\Delta_\tau'\) then
\[
  \bigl\|\mathrm{TE}(\tau_2)-\mathrm{TE}(\tau_1)\bigr\|
  \;=\;
  \bigl\|\mathrm{TE}(\tau'_2)-\mathrm{TE}(\tau'_1)\bigr\|.
\]
That is, the Euclidean distance between two sinusoidal embeddings
depends \emph{only} on the time difference, not on the absolute
timestamps.
\end{theorem}

\begin{proof}[Step-by-step proof]

\smallskip
\hrule\smallskip
\noindent

Fix an index \(n\in\{0,\dots,\tfrac{d}{2}-1\}\) and two times
\(t_1,t_2\in\mathbb{N}\).  Denote
\[
  \Delta = t_2-t_1,
  \qquad
  \Sigma = t_1+t_2.
\]
Using the standard sum-and-difference identities,
\begin{align*}
  \sin(\omega_n t_2)-\sin(\omega_n t_1)
  &= 2\,
     \sin\!\Bigl(\tfrac{\omega_n\Delta}{2}\Bigr)\,
     \cos\!\Bigl(\tfrac{\omega_n\Sigma}{2}\Bigr), \\[6pt]
  \cos(\omega_n t_2)-\cos(\omega_n t_1)
  &= -\,2\,
     \sin\!\Bigl(\tfrac{\omega_n\Delta}{2}\Bigr)\,
     \sin\!\Bigl(\tfrac{\omega_n\Sigma}{2}\Bigr).
\end{align*}

Thus, the squared contribution of this frequency to the
Euclidean distance is
\[
  \bigl[\sin(\omega_n t_2)-\sin(\omega_n t_1)\bigr]^2
  + \bigl[\cos(\omega_n t_2)-\cos(\omega_n t_1)\bigr]^2
  \;=\;
  4\,\sin^2\!\Bigl(\tfrac{\omega_n\Delta}{2}\Bigr),
\]
because \(\cos^2+\sin^2=1\) removes any dependence on
\(\Sigma\).

Summing over all \(n\) gives
\[
  \bigl\|\mathrm{TE}(t_2)-\mathrm{TE}(t_1)\bigr\|^2
  \;=\;
  \sum_{n=0}^{\tfrac{d}{2}-1}
    4\,\sin^2\!\Bigl(\tfrac{\omega_n\Delta}{2}\Bigr)
  \;=\;
  4\sum_{n=0}^{\tfrac{d}{2}-1}
    \sin^2\!\Bigl(\tfrac{\omega_n\Delta}{2}\Bigr),
\]
which is a function of \(\Delta=t_2-t_1\) alone.
Consequently, if \(\tau_2-\tau_1=\tau'_2-\tau'_1\) then
\(\Delta_\tau=\Delta_\tau'\) and
\[
  \bigl\|\mathrm{TE}(\tau_2)-\mathrm{TE}(\tau_1)\bigr\|^2
  = 4\sum_{n}
      \sin^2\!\Bigl(\tfrac{\omega_n\Delta_\tau}{2}\Bigr)
  = 4\sum_{n}
      \sin^2\!\Bigl(\tfrac{\omega_n\Delta_\tau'}{2}\Bigr)
  = \bigl\|\mathrm{TE}(\tau'_2)-\mathrm{TE}(\tau'_1)\bigr\|^2.
\]
Taking square roots results in the claimed equality of norms.
\end{proof}

\paragraph{Capturing Diverse Periodicities}
Temporal facts expressed as $(s, p, o, \tau)$, where $\tau \in \mathbb{T}$ represents the time annotation, can capture periodic phenomena, such as the Olympic Games occurring every four years or annual events like the Nobel Prize ceremonies.
Modeling diverse periodic 
behaviors is crucial in the zero-shot transfer learning, as it enables systems to accurately represent and analyze the dynamic evolution of diverse relationships and events. 
In this section, we prove that \ModelName{} is capable of capturing periodic (with different frequencies) 
temporal facts.   
In the following theorem, we prove that the scorer in \autoref{scorefunctionequation} can capture and express various frequencies even if we assume there $f_{\theta}$ is simply a linear function (with $m$ linear nodes for representing $m$ different frequencies). For simplicity, we loosely use $\tau$ as the timestamp index $i$ in $\tau_i$ in the following.

\begin{theorem}[Multi-Frequency Affine Scorer
]
\label{thm:multi-frequency}
Let
\begin{itemize}
    \item $T>1$ be an integer time horizon ;
    \item $d_{\mathrm{PE}}\ge T$ be an even positional–encoding dimension;
    \item $P_1,\dots ,P_m,$ such that $P_i\mid T$ for every $i=1,\dots ,m,$ is a family of positive integers.
    \item $[\mathrm{TE}(\tau)]_{2n}   = \sin(\omega_n\tau),
            \qquad
            [\mathrm{TE}(\tau)]_{2n+1} = \cos(\omega_n\tau),
            \qquad
            0\le n < \tfrac{d_{\mathrm{PE}}}{2},$
             with arbitrary real frequencies $\omega_0,\dots ,\omega_{d_{\mathrm{PE}}/2-1}$, be (standard) sinusoidal positional encoding defined for any $\tau$.

    \item $G = \{V_1,\dots,V_m\}\subset\mathbb{R}^{d_V}$ be $m$ fixed context vectors and,
    \item  $g_i : \{0,1,\dots ,T-1\}\longrightarrow \mathbb{R}$
 be a \emph{non-constant} sequence for each $i=1,\dots ,m$, that we extend to all $\mathbb{Z}$ by $P_i$-periodicity:
\[
  g_i(\tau) := g_i\!\bigl(\tau\bmod P_i\bigr),
  \qquad \tau\in\mathbb{Z}.
\]
    \item and \[
  M \;=\;
  \begin{pmatrix}
    \mathrm{TE}(0)^\top\\
    \mathrm{TE}(1)^\top\\
    \vdots\\
    \mathrm{TE}(T-1)^\top
  \end{pmatrix}
  \in\mathbb{R}^{T\times d_{\mathrm{PE}}},
  \qquad
  B^{(i)}
  =
  \begin{pmatrix}
    \mathrm{TE}(P_i)^\top    - \mathrm{TE}(0)^\top\\
    \mathrm{TE}(P_i+1)^\top  - \mathrm{TE}(1)^\top\\
    \vdots\\
    \mathrm{TE}(T-1+P_i)^\top- \mathrm{TE}(T-1)^\top
  \end{pmatrix}
  \in\mathbb{R}^{T\times d_{\mathrm{PE}}}.
\] be encoding matrices, we stack $B^{(i)}$ block-diagonally:
\[
  \widetilde B := \operatorname{diag}\!\bigl(B^{(1)},\dots ,B^{(m)}\bigr)
  \in\mathbb{R}^{mT\times m\,d_{\mathrm{PE}}}.
\]

\end{itemize}

\medskip
\noindent

\medskip
\noindent

\medskip
\noindent

\bigskip
\noindent
Define
\[
  g:=
  \begin{pmatrix}
    g_1\\
    \vdots\\
    g_m
  \end{pmatrix}
  \in\mathbb{R}^{mT},
  \qquad
  w_{\mathrm{PE}}:=
  \begin{pmatrix}
    w_{\mathrm{PE}}^{(1)}\\
    \vdots\\
    w_{\mathrm{PE}}^{(m)}
  \end{pmatrix}
  \in\mathbb{R}^{m\,d_{\mathrm{PE}}}.
\]

\begin{enumerate}[label=\textup{(\Alph*)},leftmargin=*,itemsep=4pt]
  \item \textbf{Compatibility \((\mathcal{C})\).}\;  
        There exists $w_{\mathrm{PE}}\in\mathbb{R}^{m\,d_{\mathrm{PE}}}$ solving the linear system
        \[
          \boxed{\;
          \begin{pmatrix}
            I_m\!\otimes\! M \\[4pt]
            \widetilde B
          \end{pmatrix}
          w_{\mathrm{PE}}
          =
          \begin{pmatrix}
            g\\[3pt]0
          \end{pmatrix}}
          \tag{$\mathcal{C}$}
        \]
        where $I_m\otimes M$ is the block-diagonal Kronecker product whose $i$-th diagonal block equals $M$.
  \item \textbf{Existence of a scorer \((\mathcal{S})\).}\;  
        There exist parameters
        \[
          \theta
          \;=\;
          \bigl(W_{\mathrm{PE}},\,w_V,\,b\bigr)
          \in
          \mathbb{R}^{d_{\mathrm{PE}}\times m}
          \times\mathbb{R}^{d_V}
          \times\mathbb{R}
        \]
        such that the affine map
        \[
          f_\theta\bigl(V,\mathrm{TE}(\tau)\bigr)
          :=
          w_V^\top V
          \;+\;
          W_{\mathrm{PE}}^\top\,\mathrm{TE}(\tau)
          \;+\;
          b
          \;\in\mathbb{R}^{m}
        \]
        obeys, for each $i=1,\dots ,m$,
        \begin{enumerate}[label=\textup{(\roman*)}]
          \item $P_i$-periodicity in $\tau\in\mathbb{Z}$;
          \item non-constancy as a function of $\tau$;
          \item interpolation on the basic window:
                \[
                  \bigl[f_\theta\!\bigl(V_i,\mathrm{TE}(\tau)\bigr)\bigr]_i
                  \;=\;
                  w_V^\top V_i + g_i(\tau),
                  \qquad
                  0\le\tau<T.
                \]
        \end{enumerate}
\end{enumerate}

Given $\mathcal{C}$ and $\mathcal{S}$, the following holds
\begin{center}
\framebox{\(\mathcal{C}\;\Longleftrightarrow\;\mathcal{S}\)}.
\end{center}

\bigskip
\noindent
When these equivalent conditions hold one may choose \emph{any} solution $w_{\mathrm{PE}}$ of~\((\mathcal{C})\), set
\[
  W_{\mathrm{PE}}
  :=
  \bigl[w_{\mathrm{PE}}^{(1)}\;\cdots\;w_{\mathrm{PE}}^{(m)}\bigr],
  \qquad
  b=0,
\]
and pick an arbitrary $w_V$; the resulting $\theta$ satisfies \(\mathcal{S}\).
\end{theorem}

\begin{proof}[Step-by-step proof of \(\mathcal{C}\Longleftrightarrow\mathcal{S}\)]
We write $w_{\mathrm{PE}}=(w_{\mathrm{PE}}^{(1)},\dots ,w_{\mathrm{PE}}^{(m)})$ with
each block $w_{\mathrm{PE}}^{(i)}\in\mathbb{R}^{d_{\mathrm{PE}}}$.

\smallskip
\hrule\smallskip
\noindent
\textbf{Part I: \(\mathcal{C}\;\Longrightarrow\;\mathcal{S}\).}
Assume a vector $w_{\mathrm{PE}}$ satisfies system~\((\mathcal{C})\). 
Here, we show that if the compatibility condition holds, then a scorer exists. We divide the proof into four parts as follows:

\begin{enumerate}[label=\textbf{Step~\arabic*:},leftmargin=*,itemsep=6pt]
\item 
      The top block $(I_m\otimes M)w_{\mathrm{PE}}=g$ is equivalent to
      \[
        M\,w_{\mathrm{PE}}^{(i)} \;=\; g_i \Big|_{\,0\le\tau<T},
        \quad
        i=1,\dots ,m.
      \tag{1}
      \]
\item 
      The block $\widetilde B\,w_{\mathrm{PE}}=0$ gives for every $i$
      \[
        B^{(i)}\,w_{\mathrm{PE}}^{(i)} = 0
        \;\Longrightarrow\;
        w_{\mathrm{PE}}^{(i)\top}
        \bigl(\mathrm{TE}(\tau+P_i)-\mathrm{TE}(\tau)\bigr)=0,
        \quad
        0\le\tau<T.
      \]
      Thus
      \[
        w_{\mathrm{PE}}^{(i)\top}\mathrm{TE}(\tau+P_i)
        =w_{\mathrm{PE}}^{(i)\top}\mathrm{TE}(\tau),
        \quad 0\le\tau<T.
        \tag{2}
      \]
      Replacing $\tau$ by $\tau+P_i$ and iterating $k$ times shows
      \[
        w_{\mathrm{PE}}^{(i)\top}\mathrm{TE}(\tau+kP_i)
        = w_{\mathrm{PE}}^{(i)\top}\mathrm{TE}(\tau),
        \quad
        \forall\tau\in\{0,\dots ,T-1\},\;k\in\mathbb{Z},
      \]
      i.e., $P_i$-periodicity on \emph{all} integers.
\item 
      Combine~(1) with periodicity: for every $\tau\in\mathbb{Z}$
      \[
        w_{\mathrm{PE}}^{(i)\top}\mathrm{TE}(\tau)
        = g_i\!\bigl(\tau\bmod P_i\bigr).
        \tag{3}
      \]
\item 
      Define
      \[
        W_{\mathrm{PE}}
        :=
        \bigl[w_{\mathrm{PE}}^{(1)}\;\cdots\;w_{\mathrm{PE}}^{(m)}\bigr],
        \qquad
        b:=0,
        \qquad
        \text{choose any }w_V\in\mathbb{R}^{d_V}.
      \]
      Then
      \[
        f_\theta\bigl(V,\mathrm{TE}(\tau)\bigr)
        =
        w_V^\top V
        +
        \begin{pmatrix}
          w_{\mathrm{PE}}^{(1)\top}\mathrm{TE}(\tau)\\[-1pt]
          \vdots\\[-1pt]
          w_{\mathrm{PE}}^{(m)\top}\mathrm{TE}(\tau)
        \end{pmatrix},
      \]
      whose $i$-th component equals the right-hand side of~(3).  
      Therefore:

      a) it is $P_i$-periodic by construction;  
      b) it is non-constant because each $g_i$ is non-constant;  
      c) on $0\le\tau<T$, (3) reduces to \(w_{\!V}^{\top}V_i+g_i(\tau)\).

      Hence \(\mathcal{S}\) holds.
\end{enumerate}

\smallskip
\hrule\smallskip
\noindent
\textbf{Part II: \(\mathcal{S}\;\Longrightarrow\;\mathcal{C}\).}

Conversely, suppose parameters \(\theta=(W_{\mathrm{PE}},w_V,b)\) satisfy \(\mathcal{S}\).
Write the columns as
\(
  W_{\mathrm{PE}}=[\,w_{\mathrm{PE}}^{(1)}\mid\cdots\mid w_{\mathrm{PE}}^{(m)}].
\)

\begin{enumerate}[label=\textbf{Step~\arabic*:},leftmargin=*,itemsep=6pt]
\item 
      Evaluating property~(iii) at $0\le\tau<T$ gives
      \[
        M\,w_{\mathrm{PE}}^{(i)} = g_i \Big|_{\,0\le\tau<T},
        \quad
        i=1,\dots ,m.
        \tag{4}
      \]
\item \emph{$P_i$-periodicity implies the $B^{(i)}$ equations.}\;
      Property (i) yields, for every integer $\tau$,
      \(
        w_{\mathrm{PE}}^{(i)\top}\mathrm{TE}(\tau+P_i)=
        w_{\mathrm{PE}}^{(i)\top}\mathrm{TE}(\tau)
      \).
      Specialising to $\tau=0,\dots ,T-1$ one obtains
      \(
        B^{(i)}\,w_{\mathrm{PE}}^{(i)}=0
      \).
\item 
      Collecting (4) for all $i$ and the $B^{(i)}$ equations in a single vector $w_{\mathrm{PE}}$ yields exactly system~\((\mathcal{C})\).  Thus \(\mathcal{C}\) holds.
\end{enumerate}

\smallskip
\hrule\smallskip
\noindent
Both directions are proven; therefore \(\mathcal{C}\Longleftrightarrow\mathcal{S}\).  
The final comment about choosing $(W_{\mathrm{PE}},w_V,b)$ is precisely the construction in Part I, Step 4.
\end{proof}

Following \autoref{thm:multi-frequency}, we are interested in knowing under which conditions the compatibility condition holds itself, i.e., when the system has a solution.

\begin{theorem}[Universal Compatibility under harmonically–aligned frequencies]%
\label{thm:compat-always}
Fix
\[
  T>1,\qquad d_{\mathrm{PE}}\ge T,\qquad
  P_1,\dots ,P_m\in\mathbb N,\;\;P_i\mid T.
\]

\medskip
\noindent
Let \(L:=\operatorname{lcm}(P_1,\dots ,P_m)\)\footnote{least common multiple} and choose \emph{any} collection of
integers \(k_0,k_1,\dots ,k_{d_{\mathrm{PE}}/2-1}\).
Define the frequencies
\[
  \omega_n \;:=\; \frac{2\pi k_n}{L},
  \qquad 0\le n<\tfrac{d_{\mathrm{PE}}}{2},
\]
and build the sinusoidal positional encoding
\[
  [\mathrm{TE}(\tau)]_{2n}   = \sin(\omega_n\tau),
  \quad
  [\mathrm{TE}(\tau)]_{2n+1} = \cos(\omega_n\tau),
  \qquad \tau\in\mathbb N.
\]

Form the matrices as follows
\[
  M\;=\;
  \begin{pmatrix}\mathrm{TE}(0)^\top\\ \vdots\\ \mathrm{TE}(T-1)^\top\end{pmatrix}\in\mathbb R^{T\times d_{\mathrm{PE}}},
  \qquad
  B^{(i)}
  =
  \begin{pmatrix}
    \mathrm{TE}(P_i)^\top-\mathrm{TE}(0)^\top\\
    \vdots\\
    \mathrm{TE}(T-1+P_i)^\top-\mathrm{TE}(T-1)^\top
  \end{pmatrix}
  \in\mathbb R^{T\times d_{\mathrm{PE}}}.
\]

For these frequencies, every \(B^{(i)}\) vanishes identically; hence, the stacked
compatibility system
\[
  \begin{pmatrix}
    I_m\otimes M \\[4pt] \mathrm{diag}(B^{(1)},\dots ,B^{(m)})
  \end{pmatrix}
  w_{\mathrm{PE}}
  =
  \begin{pmatrix}
    g\\[3pt]0
  \end{pmatrix}
  \tag{$\mathcal{C}$}
\]
is always solvable, regardless of the choice of \emph{any} target sequences
\(g_i\colon\{0,\dots ,T-1\}\to\mathbb R\).
\end{theorem}

\smallskip
\hrule\smallskip
\noindent

\begin{proof}[We present the Step-by-step proof as follows.]

\textbf{Step 1.  Encoding is \(L\)-periodic.}
Because \(\omega_n L=2\pi k_n\),
\[
  \sin(\omega_n(\tau+L))=\sin(\omega_n\tau),\quad
  \cos(\omega_n(\tau+L))=\cos(\omega_n\tau),
  \qquad\forall\tau\in\mathbb Z,
\]
so \(\mathrm{TE}(\tau+L)=\mathrm{TE}(\tau)\).

\smallskip
\textbf{Step 2.  Encoding is also \(P_i\)-periodic.}
Each \(P_i\) divides \(L\), hence \(\mathrm{TE}(\tau+P_i)=\mathrm{TE}(\tau)\).
Therefore
\[
  B^{(i)}=0\in\mathbb R^{T\times d_{\mathrm{PE}}},
  \qquad i=1,\dots ,m.
\]

\smallskip
\textbf{Step 3.  Compatibility degenerates to a single block.}
With every \(B^{(i)}\) equal to zero, system \((\mathcal{C})\) becomes
\[
  (I_m\otimes M)\,w_{\mathrm{PE}} \;=\; g .
  \tag{$\mathcal{C}'$}
\]

\smallskip
\textbf{Step 4.  Solve the reduced system.}
Write \(w_{\mathrm{PE}}=(w_{\mathrm{PE}}^{(1)},\dots ,w_{\mathrm{PE}}^{(m)})\)
and \(g=(g_1,\dots ,g_m)\) with blocks in \(\mathbb R^{T}\).
The Kronecker structure of \(I_m\otimes M\) splits \((\mathcal{C}')\) into
\(m\) independent systems
\[
  M\,w_{\mathrm{PE}}^{(i)} = g_i,\qquad i=1,\dots ,m.
  \tag{$\mathcal{C}_i$}
\]
Because \(d_{\mathrm{PE}}\ge T\), the rows of \(M\) are linearly dependent
at worst; pick any right inverse \(M^{+}\) (for instance the Moore–Penrose
pseudoinverse).  Then
\[
  w_{\mathrm{PE}}^{(i)} := M^{+}g_i
  \quad (i=1,\dots ,m)
\]
solves each \((\mathcal{C}_i)\), and hence \(w_{\mathrm{PE}}\) solves the full
system \((\mathcal{C})\).

\smallskip
\textbf{Step 5.  Conclusion.}
Compatibility holds \emph{for every choice of data} \(\{g_i\}\)
once the frequencies obey \(\omega_n=\tfrac{2\pi k_n}{L}\).
\end{proof}

%% file: neurips_2025.bbl
\begin{thebibliography}{40}
\providecommand{\natexlab}[1]{#1}
\providecommand{\url}[1]{\texttt{#1}}
\expandafter\ifx\csname urlstyle\endcsname\relax
  \providecommand{\doi}[1]{doi: #1}\else
  \providecommand{\doi}{doi: \begingroup \urlstyle{rm}\Url}\fi

\bibitem[Ao et~al.(2022)Ao, Doyle, Healey, and Vatsavai]{10.5555/AAI30283568}
Jing Ao, Jon Doyle, Christopher Healey, and Ranga~Raju Vatsavai.
\newblock \emph{Temporal Knowledge Graphs: Integration, Querying, and Analytics}.
\newblock PhD thesis, North Carolina State University, 2022.
\newblock AAI30283568.

\bibitem[Bordes et~al.(2013)Bordes, Usunier, Garcia-Duran, Weston, and Yakhnenko]{bordes2013translating}
Antoine Bordes, Nicolas Usunier, Alberto Garcia-Duran, Jason Weston, and Oksana Yakhnenko.
\newblock Translating embeddings for modeling multi-relational data.
\newblock \emph{Advances in neural information processing systems}, 26, 2013.

\bibitem[Cai et~al.(2024)Cai, Mao, Zhou, Long, Wu, and Lan]{cai2024surveytemporalknowledgegraph}
Li~Cai, Xin Mao, Yuhao Zhou, Zhaoguang Long, Changxu Wu, and Man Lan.
\newblock A survey on temporal knowledge graph: Representation learning and applications, 2024.
\newblock URL \url{https://arxiv.org/abs/2403.04782}.

\bibitem[Chen et~al.(2024)Chen, Wang, Li, Li, Yu, and Song]{chen2024unified}
Kai Chen, Ye~Wang, Yitong Li, Aiping Li, Han Yu, and Xin Song.
\newblock A unified temporal knowledge graph reasoning model towards interpolation and extrapolation.
\newblock In \emph{Proceedings of the 62nd Annual Meeting of the Association for Computational Linguistics (Volume 1: Long Papers)}, pages 117--132, 2024.

\bibitem[Chen et~al.(2022)Chen, Zhang, Zhu, Zhou, Yuan, Xu, and Chen]{chen2022meta}
Mingyang Chen, Wen Zhang, Yushan Zhu, Hongting Zhou, Zonggang Yuan, Changliang Xu, and Huajun Chen.
\newblock Meta-knowledge transfer for inductive knowledge graph embedding.
\newblock In \emph{Proceedings of the 45th international ACM SIGIR conference on research and development in information retrieval}, pages 927--937, 2022.

\bibitem[Chen et~al.(2023)Chen, Xu, Su, Huang, and Dou]{chen2023incorporating}
Zhongwu Chen, Chengjin Xu, Fenglong Su, Zhen Huang, and Yong Dou.
\newblock Incorporating structured sentences with time-enhanced bert for fully-inductive temporal relation prediction.
\newblock In \emph{Proceedings of the 46th International ACM SIGIR Conference on Research and Development in Information Retrieval}, pages 889--899, 2023.

\bibitem[Ding et~al.(2025)Ding, Huang, Yu, Jin, and He]{ding2025towards}
Ling Ding, Lei Huang, Zhizhi Yu, Di~Jin, and Dongxiao He.
\newblock Towards global-topology relation graph for inductive knowledge graph completion.
\newblock In \emph{Proceedings of the AAAI Conference on Artificial Intelligence}, volume~39, pages 11581--11589, 2025.

\bibitem[Ding et~al.(2024)Ding, Cai, Wu, Ma, Liao, Xiong, and Tresp]{ding2024zrllm}
Zifeng Ding, Heling Cai, Jingpei Wu, Yunpu Ma, Ruotong Liao, Bo~Xiong, and Volker Tresp.
\newblock zrllm: Zero-shot relational learning on temporal knowledge graphs with large language models.
\newblock In \emph{Proceedings of the 2024 Conference of the North American Chapter of the Association for Computational Linguistics: Human Language Technologies (Volume 1: Long Papers)}, pages 1877--1895, 2024.

\bibitem[Galkin et~al.(2024)Galkin, Yuan, Mostafa, Tang, and Zhu]{galkintowards}
Mikhail Galkin, Xinyu Yuan, Hesham Mostafa, Jian Tang, and Zhaocheng Zhu.
\newblock Towards foundation models for knowledge graph reasoning.
\newblock In \emph{The Twelfth International Conference on Learning Representations}, 2024.

\bibitem[Garc{\'i}a-Dur{\'a}n et~al.(2018)Garc{\'i}a-Dur{\'a}n, Duman{\v{c}}i{\'c}, and Niepert]{garcia-duran-etal-2018-learning}
Alberto Garc{\'i}a-Dur{\'a}n, Sebastijan Duman{\v{c}}i{\'c}, and Mathias Niepert.
\newblock Learning sequence encoders for temporal knowledge graph completion.
\newblock In Ellen Riloff, David Chiang, Julia Hockenmaier, and Jun{'}ichi Tsujii, editors, \emph{Proceedings of the 2018 Conference on Empirical Methods in Natural Language Processing}, pages 4816--4821, Brussels, Belgium, October-November 2018. Association for Computational Linguistics.
\newblock \doi{10.18653/v1/D18-1516}.
\newblock URL \url{https://aclanthology.org/D18-1516/}.

\bibitem[Gastinger et~al.(2023)Gastinger, Sztyler, Sharma, Schuelke, and Stuckenschmidt]{gastinger2023comparing}
Julia Gastinger, Timo Sztyler, Lokesh Sharma, Anett Schuelke, and Heiner Stuckenschmidt.
\newblock Comparing apples and oranges? on the evaluation of methods for temporal knowledge graph forecasting.
\newblock In \emph{Joint European conference on machine learning and knowledge discovery in databases}, pages 533--549. Springer, 2023.

\bibitem[Goel et~al.(2020)Goel, Kazemi, Brubaker, and Poupart]{goel2020diachronic}
Rishab Goel, Seyed~Mehran Kazemi, Marcus Brubaker, and Pascal Poupart.
\newblock Diachronic embedding for temporal knowledge graph completion.
\newblock In \emph{Proceedings of the AAAI Conference on Artificial Intelligence}, volume~34, pages 3988--3995, 2020.

\bibitem[Han et~al.(2020)Han, Chen, Ma, and Tresp]{han2020explainable}
Zhen Han, Peng Chen, Yunpu Ma, and Volker Tresp.
\newblock Explainable subgraph reasoning for forecasting on temporal knowledge graphs.
\newblock In \emph{International conference on learning representations}, 2020.

\bibitem[Han et~al.(2023)Han, Liao, Gu, Zhang, Ding, Gu, Koeppl, Sch{\"u}tze, and Tresp]{han2023ecola}
Zhen Han, Ruotong Liao, Jindong Gu, Yao Zhang, Zifeng Ding, Yujia Gu, Heinz Koeppl, Hinrich Sch{\"u}tze, and Volker Tresp.
\newblock Ecola: Enhancing temporal knowledge embeddings with contextualized language representations.
\newblock In \emph{Findings of the Association for Computational Linguistics: ACL 2023}, pages 5433--5447, 2023.

\bibitem[Jin et~al.(2020)Jin, Qu, Jin, and Ren]{jin2020recurrent}
Woojeong Jin, Meng Qu, Xisen Jin, and Xiang Ren.
\newblock Recurrent event network: Autoregressive structure inferenceover temporal knowledge graphs.
\newblock In \emph{Proceedings of the 2020 Conference on Empirical Methods in Natural Language Processing (EMNLP)}, pages 6669--6683, 2020.

\bibitem[Lacroix et~al.(2020)Lacroix, Obozinski, and Usunier]{tcomplexlacroix2020tensor}
Timothée Lacroix, Guillaume Obozinski, and Nicolas Usunier.
\newblock Tensor decompositions for temporal knowledge base completion.
\newblock In \emph{International Conference on Learning Representations}, 2020.
\newblock URL \url{https://openreview.net/forum?id=rke2P1BFwS}.

\bibitem[Lautenschlager et~al.(2015)Lautenschlager, Shellman, and Ward]{lautenschlager2015icews}
Jennifer Lautenschlager, Steve Shellman, and Michael Ward.
\newblock Icews event aggregations.
\newblock \emph{Harvard Dataverse}, 3\penalty0 (595):\penalty0 28, 2015.

\bibitem[Leblay and Chekol(2018)]{leblay2018deriving}
Julien Leblay and Melisachew~Wudage Chekol.
\newblock Deriving validity time in knowledge graph.
\newblock In \emph{Companion proceedings of the the web conference 2018}, pages 1771--1776, 2018.

\bibitem[Lee et~al.(2023{\natexlab{a}})Lee, Ahrabian, Jin, Morstatter, and Pujara]{lee2023temporal}
Dong-Ho Lee, Kian Ahrabian, Woojeong Jin, Fred Morstatter, and Jay Pujara.
\newblock Temporal knowledge graph forecasting without knowledge using in-context learning.
\newblock In \emph{Proceedings of the 2023 Conference on Empirical Methods in Natural Language Processing}, pages 544--557, 2023{\natexlab{a}}.

\bibitem[Lee et~al.(2023{\natexlab{b}})Lee, Chung, and Whang]{lee2023ingram}
Jaejun Lee, Chanyoung Chung, and Joyce~Jiyoung Whang.
\newblock Ingram: Inductive knowledge graph embedding via relation graphs.
\newblock In \emph{International Conference on Machine Learning}, pages 18796--18809. PMLR, 2023{\natexlab{b}}.

\bibitem[Li et~al.(2022{\natexlab{a}})Li, Sun, and Zhao]{li2022tirgn}
Yujia Li, Shiliang Sun, and Jing Zhao.
\newblock Tirgn: Time-guided recurrent graph network with local-global historical patterns for temporal knowledge graph reasoning.
\newblock In \emph{IJCAI}, pages 2152--2158, 2022{\natexlab{a}}.

\bibitem[Li et~al.(2022{\natexlab{b}})Li, Guan, Jin, Peng, Lyu, Zhu, Bai, Li, Guo, and Cheng]{li2022complex}
Zixuan Li, Saiping Guan, Xiaolong Jin, Weihua Peng, Yajuan Lyu, Yong Zhu, Long Bai, Wei Li, Jiafeng Guo, and Xueqi Cheng.
\newblock Complex evolutional pattern learning for temporal knowledge graph reasoning.
\newblock In \emph{Proceedings of the 60th Annual Meeting of the Association for Computational Linguistics (Volume 2: Short Papers)}, pages 290--296, 2022{\natexlab{b}}.

\bibitem[Liang et~al.(2023)Liang, Meng, Liu, Liu, Tu, Wang, Zhou, and Liu]{liang2023learn}
Ke~Liang, Lingyuan Meng, Meng Liu, Yue Liu, Wenxuan Tu, Siwei Wang, Sihang Zhou, and Xinwang Liu.
\newblock Learn from relational correlations and periodic events for temporal knowledge graph reasoning.
\newblock In \emph{Proceedings of the 46th international ACM SIGIR conference on research and development in information retrieval}, pages 1559--1568, 2023.

\bibitem[Liao et~al.(2024)Liao, Jia, Li, Ma, and Tresp]{liao2024gentkg}
Ruotong Liao, Xu~Jia, Yangzhe Li, Yunpu Ma, and Volker Tresp.
\newblock Gentkg: Generative forecasting on temporal knowledge graph with large language models.
\newblock In \emph{NAACL-HLT (Findings)}, 2024.

\bibitem[Liu et~al.(2021)Liu, Grau, Horrocks, and Kostylev]{liu2021indigo}
Shuwen Liu, Bernardo Grau, Ian Horrocks, and Egor Kostylev.
\newblock Indigo: Gnn-based inductive knowledge graph completion using pair-wise encoding.
\newblock \emph{Advances in Neural Information Processing Systems}, 34:\penalty0 2034--2045, 2021.

\bibitem[Liu et~al.(2022)Liu, Ma, Hildebrandt, Joblin, and Tresp]{liu2022tlogic}
Yushan Liu, Yunpu Ma, Marcel Hildebrandt, Mitchell Joblin, and Volker Tresp.
\newblock Tlogic: Temporal logical rules for explainable link forecasting on temporal knowledge graphs.
\newblock In \emph{Proceedings of the AAAI conference on artificial intelligence}, volume~36, pages 4120--4127, 2022.

\bibitem[Pan et~al.(2024)Pan, Nayyeri, Li, and Staab]{pan2024hge}
Jiaxin Pan, Mojtaba Nayyeri, Yinan Li, and Steffen Staab.
\newblock Hge: embedding temporal knowledge graphs in a product space of heterogeneous geometric subspaces.
\newblock In \emph{Proceedings of the AAAI Conference on Artificial Intelligence}, volume~38, pages 8913--8920, 2024.

\bibitem[Saxena et~al.(2021)Saxena, Chakrabarti, and Talukdar]{saxena-etal-2021-question}
Apoorv Saxena, Soumen Chakrabarti, and Partha Talukdar.
\newblock Question answering over temporal knowledge graphs.
\newblock In Chengqing Zong, Fei Xia, Wenjie Li, and Roberto Navigli, editors, \emph{Proceedings of the 59th Annual Meeting of the Association for Computational Linguistics and the 11th International Joint Conference on Natural Language Processing (Volume 1: Long Papers)}, pages 6663--6676, Online, August 2021. Association for Computational Linguistics.
\newblock \doi{10.18653/v1/2021.acl-long.520}.
\newblock URL \url{https://aclanthology.org/2021.acl-long.520/}.

\bibitem[Su et~al.(2024)Su, Ahmed, Lu, Pan, Bo, and Liu]{su2024roformer}
Jianlin Su, Murtadha Ahmed, Yu~Lu, Shengfeng Pan, Wen Bo, and Yunfeng Liu.
\newblock Roformer: Enhanced transformer with rotary position embedding.
\newblock \emph{Neurocomputing}, 568:\penalty0 127063, 2024.

\bibitem[Sun et~al.(2021)Sun, Zhong, Ma, Han, and He]{sun2021timetraveler}
Haohai Sun, Jialun Zhong, Yunpu Ma, Zhen Han, and Kun He.
\newblock Timetraveler: Reinforcement learning for temporal knowledge graph forecasting.
\newblock In \emph{Proceedings of the 2021 Conference on Empirical Methods in Natural Language Processing}, pages 8306--8319, 2021.

\bibitem[Sun et~al.(2025)Sun, Huang, Zhou, Wan, Peng, and Yu]{sun2025riemanngfm}
Li~Sun, Zhenhao Huang, Suyang Zhou, Qiqi Wan, Hao Peng, and Philip~S Yu.
\newblock Riemanngfm: Learning a graph foundation model from structural geometry.
\newblock In \emph{THE WEB CONFERENCE 2025}, 2025.

\bibitem[Teru et~al.(2020)Teru, Denis, and Hamilton]{teru2020inductive}
Komal Teru, Etienne Denis, and Will Hamilton.
\newblock Inductive relation prediction by subgraph reasoning.
\newblock In \emph{International conference on machine learning}, pages 9448--9457. PMLR, 2020.

\bibitem[Trivedi et~al.(2017)Trivedi, Dai, Wang, and Song]{trivedi2017know}
Rakshit Trivedi, Hanjun Dai, Yichen Wang, and Le~Song.
\newblock Know-evolve: Deep temporal reasoning for dynamic knowledge graphs.
\newblock In \emph{international conference on machine learning}, pages 3462--3471. PMLR, 2017.

\bibitem[Vaswani et~al.(2017)Vaswani, Shazeer, Parmar, Uszkoreit, Jones, Gomez, Kaiser, and Polosukhin]{vaswani2017attention}
Ashish Vaswani, Noam Shazeer, Niki Parmar, Jakob Uszkoreit, Llion Jones, Aidan~N Gomez, {\L}ukasz Kaiser, and Illia Polosukhin.
\newblock Attention is all you need.
\newblock \emph{Advances in neural information processing systems}, 30, 2017.

\bibitem[Wang et~al.(2024)Wang, Zhang, Chawla, Zhang, and Ye]{wang2024gft}
Zehong Wang, Zheyuan Zhang, Nitesh Chawla, Chuxu Zhang, and Yanfang Ye.
\newblock Gft: Graph foundation model with transferable tree vocabulary.
\newblock \emph{Advances in Neural Information Processing Systems}, 37:\penalty0 107403--107443, 2024.

\bibitem[Xu et~al.(2020)Xu, Nayyeri, Alkhoury, Yazdi, and Lehmann]{xu2020tero}
Chengjin Xu, Mojtaba Nayyeri, Fouad Alkhoury, Hamed~Shariat Yazdi, and Jens Lehmann.
\newblock Tero: A time-aware knowledge graph embedding via temporal rotation.
\newblock In \emph{Proceedings of the 28th International Conference on Computational Linguistics}, pages 1583--1593, 2020.

\bibitem[Xu et~al.(2023)Xu, Liu, Peng, Jia, and Peng]{xu2023pre}
Wenjie Xu, Ben Liu, Miao Peng, Xu~Jia, and Min Peng.
\newblock Pre-trained language model with prompts for temporal knowledge graph completion.
\newblock In \emph{Findings of the Association for Computational Linguistics: ACL 2023}, pages 7790--7803, 2023.

\bibitem[Yang et~al.(2015)Yang, Yih, He, Gao, and Deng]{yang2015embedding}
Bishan Yang, Scott Wen-tau Yih, Xiaodong He, Jianfeng Gao, and Li~Deng.
\newblock Embedding entities and relations for learning and inference in knowledge bases.
\newblock In \emph{Proceedings of the International Conference on Learning Representations (ICLR) 2015}, 2015.

\bibitem[Zhang et~al.(2022)Zhang, Zhang, Ao, Zhuang, Xu, and He]{tltcomplexzhang2022along}
Fuwei Zhang, Zhao Zhang, Xiang Ao, Fuzhen Zhuang, Yongjun Xu, and Qing He.
\newblock Along the time: Timeline-traced embedding for temporal knowledge graph completion.
\newblock In \emph{Proceedings of the 31st ACM International Conference on Information \& Knowledge Management}, pages 2529--2538, 2022.

\bibitem[Zhu et~al.(2021)Zhu, Zhang, Xhonneux, and Tang]{zhu2021neural}
Zhaocheng Zhu, Zuobai Zhang, Louis-Pascal Xhonneux, and Jian Tang.
\newblock Neural bellman-ford networks: A general graph neural network framework for link prediction.
\newblock \emph{Advances in neural information processing systems}, 34:\penalty0 29476--29490, 2021.

\end{thebibliography}
